%% file: main.tex
\documentclass{article} 
\usepackage{iclr2027_conference,times}

\input{math_commands.tex}

\usepackage{hyperref}
\usepackage{url}
\usepackage[textsize=scriptsize]{todonotes}
\usepackage{enumitem}
\usepackage{cleveref}
\usepackage[most]{tcolorbox}
\usepackage{float}

\setlist[itemize]{noitemsep}
\definecolor{chris}{RGB}{110,150,220}
\definecolor{promptbg}{RGB}{248,249,251}
\definecolor{promptframe}{RGB}{205,211,219}

\newcommand{\accstd}[2]{\ensuremath{#1{\scriptstyle \,\pm\, #2}}}
\newcommand{\accstdbf}[2]{\ensuremath{\textbf{#1}{\scriptstyle \,\pm\, #2}}}
\newcommand{\accstdunderline}[2]{\ensuremath{\underline{#1}{\scriptstyle \,\pm\, #2}}}

\newtcblisting{promptbox}{
  enhanced,
  breakable,
  listing only,
  colback=promptbg,
  colframe=promptframe,
  boxrule=0.4pt,
  arc=1mm,
  left=1mm,
  right=1mm,
  top=1mm,
  bottom=1mm,
  listing options={
    basicstyle=\small\ttfamily,
    breaklines=true,
    columns=fullflexible,
    keepspaces=true
  }
}

\title{Muffin - A Shared Memory Buffer for Asynchronous AI Agents}

\author{Ping-Lin Kuo, Suman Navaratnarajah, Christian Frey, Gresa Shala, Josif Grabocka \\
Department of Computer Science \& Artificial Intelligence\\
University of Technology Nuremberg\\
Dr.-Luise-Herzberg-Straße 4, 90461 Nuremberg, Germany\\
\texttt{\{firstname.lastname\}@utn.de}
}

\iclrfinalcopy 
\begin{document}

\maketitle

\fancyhead[L]{Preprint}
\pagestyle{fancy}
\thispagestyle{fancy}

\begin{abstract}
\input{Content/abstract}
\end{abstract}

\input{Content/introduction}
\input{Content/related_works}
\input{Content/03_methodology}
\input{Content/experiments}
\input{Content/discussion}
\input{Content/conclusion}

\subsection*{AI use statement}

In this work, we used generative AI tools to assist with code modifications, generate scripts for higher-quality figure plotting, and formulate prompts. We have not used generative AI tools for research ideas, designing core methodology, or writing initial text from scratch. Additionally, we used generative AI tools to improve sentence fluency while still using our original wording and arguments. We have reviewed all AI-assisted work: all modified code and plotting scripts were manually inspected, executed, and verified for correctness by the authors, and all edited text was carefully checked to ensure it had the same semantic meaning as our original version. We take responsibility for the final content of this work, including text, claims or artifacts produced with the aid of generative AI.

\subsection*{Reproducibility statement}

We provide the Muffin implementation and experiment scripts in the anonymous code repository linked in the Implementation Details paragraph in Section~\ref{subsec:setup}. Section~\ref{sec:experiments} describes the benchmark selection, dataset subsets, baselines, model choices, serving setup, repeated-run setting, and accuracy/token usage measurements. The ablation settings are also described in Section~\ref{sec:ablations}.
Appendix~\ref{app:prompts} provides the prompt templates that we use for the orchestrator, agents, summarizer, and direct-answer route. Appendix~\ref{app:additional_results} reports additional experiment results. Together, the resources mentioned above provide all settings required to reproduce the reported results.


\subsubsection*{Acknowledgments}
We acknowledge funding from the \emph{Bayerisches Staatsministerium für Wirtschaft, Landesentwicklung und Energie} under the research programme \emph{BayVFP Förderlinie Digitalisierung} (grant reference DIK-2507-0024// DIK0836/01).

We also acknowledge the funding support from the \emph{Bayerisches Landesamt fur Steuer} for the Bavarian AI Taxation Laboratory. Moreover, we gratefully acknowledge the scientific support and HPC resources provided by the Erlangen National High Performance Computing Center (NHR@FAU) of the Friedrich-Alexander-Universität Erlangen-Nürnberg (FAU) under the NHR project DeepSmall - Meta-learning for regularizing deep networks under small data regimes. NHR funding is provided by federal and Bavarian state authorities. NHR@FAU hardware is partially funded by the German Research Foundation (DFG) – 440719683.

\bibliography{iclr2027_conference}
\bibliographystyle{iclr2027_conference}

\clearpage
\appendix
\input{Content/appendix}

\end{document}

%% file: math_commands.tex
\usepackage{amsmath,amsfonts,bm}

\def\eqref#1{equation~\ref{#1}}

\def\1{\bm{1}}

\DeclareMathAlphabet{\mathsfit}{\encodingdefault}{\sfdefault}{m}{sl}
\SetMathAlphabet{\mathsfit}{bold}{\encodingdefault}{\sfdefault}{bx}{n}



%% file: Content/abstract.tex
The rise of Agentic AI systems has changed the landscape of artificial intelligence by enabling strong reasoning capabilities through specialized multi-agent interaction with structured self-reflection. However, in target-driven reasoning tasks, single- and multi-agent systems are often on par in terms of performance gains. A major challenge is that failures often arise from an ineffective exchange of critical information during inter-agent interaction and coordination. 
Different from existing multi-agent debate paradigms that rely on synchronous turn-taking and sharing information across different agents, we propose an inter-agent protocol based on shared \textbf{m}emory b\textbf{uff}er \textbf{in}ference - referred to as \textbf{Muffin} - in an asynchronous agentic framework. In the first step, an orchestrator routes queries based on their reasoning demands and delegates demanding ones to a multi-agent setting, where a shared memory buffer provides asynchronous exchange of intermediate findings while an agent's private reasoning remains local. Our experiments show that our framework Muffin outperforms established competitors, achieving the best accuracy in 17 out of 20 of our benchmark settings, and provides a better quality-cost trade-off than multi-agent debate: Muffin improves accuracy by 19.2 accuracy points while using 77.4\% fewer tokens, without requiring full-trace sharing.

%% file: Content/introduction.tex
\section{Introduction}
\label{sec:introduction}
Frontier models like Opus \citep{anthropic_2026_opus5}, Gemini \citep{google_2026_gemini38}, ChatGPT \citep{openai_2026_astra} have made a substantial leap in artificial intelligence by demonstrating reasoning capabilities that approach human performance in a variety of domains \citep{Luo_2024_NeuroscienceExperts,Bermejo_2025_TextualAnalysis,Alohali_2025_MedicalsocialSkills,street_2026_theoryofmindtasks}. 
In the current era of Agentic AI, we see a shift towards systems that are able to plan, act, use tools, and apply feedback loops for enhanced performance \citep{Abou_2025_AgenticAISurvey,Pati_2025_AgenticAI,llmbasedagentsurvey}. 
However, strong single-agent reasoning does not remove the need for collaboration. More complex tasks often require decomposing the problem, checking partial information, and integrating the conclusions generated by different reasoning parts. This motivates goal-oriented multi-agent behavior that goes beyond single-agent responses toward agent collaboration supported by memory \citep{Zhang_2025_GMemory}, reasoning \citep{subramaniam_2025_multiagentFinetuning}, or environmental interaction \citep{Lan_2024_RobotTaskPlanning}.

Recent multi-agent research has been applied to scientific experimentation \citep{Ghareeb_2026_scientificDiscovery}, embodied agents \citep{huang_2023_innerMonologue}, or societal simulation \citep{Gao_2023_socialSimulation}. 
Nevertheless, recent studies suggest that multi-agent systems do not automatically outperform strong single-agent or repeated-sampling baselines~\citep{xia_2024_agentless,kapoor2025ai}, because critical information is lost during inter-agent coordination and verification processes fail~\citep{cemri_2025_masFail}. In particular, a key limitation is that many multi-agent systems communicate after an agent has produced a complete output, leaving partial critical findings unavailable to peer agents while simultaneously overwhelming them with verbose, task-irrelevant reasoning traces that increase context overhead~\citep{liu-etal-2024-lost,ICLR2024_SoT,multiagent_debate_ICML2024}. This motivates our core question, whether agents can collaborate more effectively by asynchronously sharing the intermediate findings, and allowing other agents to retrieve and reuse information during inference without exposing full reasoning traces.

\begin{figure}[t]
    \centering
    \includegraphics[width=1\linewidth]{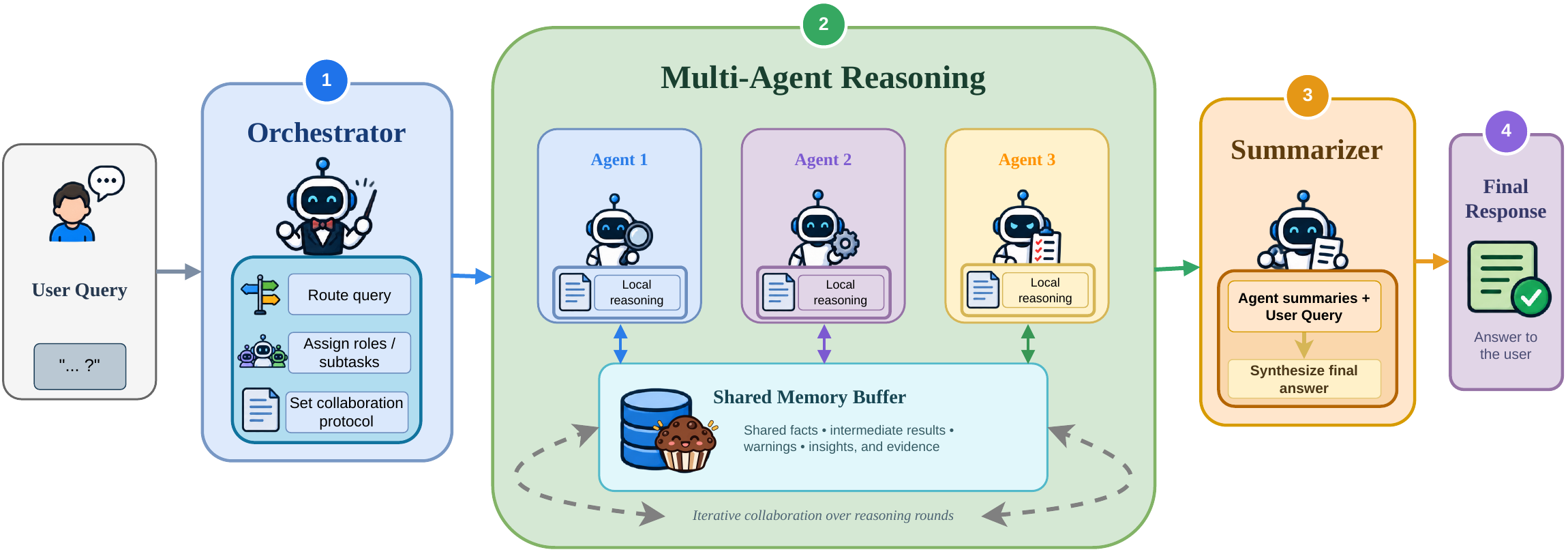}
    \caption{General workflow of Muffin - a shared memory buffering system.}
    \label{fig:overall_workflow}
\end{figure}

We propose an asynchronous system based on shared \textbf{m}emory b\textbf{uff}er \textbf{in}ference, referred to as \textbf{Muffin}, that enables exchange of agents' novel insights. The general workflow is visualized in \Cref{fig:overall_workflow}. In the first step, an \textit{Orchestrator} routes a user query based on the expected reasoning demands, i.e., whether a query can be solved directly or a multi-step reasoning is required. For more demanding queries, multiple agents are instantiated by assigning each of them a specialized role, and a subtask, both provided by the orchestrator. For an efficient exchange of relevant insights, a shared memory provides a controlled and selective exchange of intermediate findings. An agent only pushes relevant information into the buffer which is expected to help in making progress in a collaborative problem solving manner. We keep an agent's own reasoning local. After a finite reasoning horizon, a \textit{Summarizer} is fed the final messages from each agent along the agents' meta information and the initial user query, to produce the final response. 

In our empirical study, we show how Muffin provides reasoning capabilities that are on the Pareto front w.r.t. the downstream accuracy and the token consumption.
We evaluate on four reasoning benchmarks and report performance gains across four open-weight language models. By applying Muffin, we improve the strongest baseline by \textbf{4.4 accuracy points on average} across all benchmarks, with larger gains on OlymMATH-E of 11.8 accuracy points. 
Moreover, we show that it reaches a \textbf{better quality-cost trade-off} compared to multi-agent debate architectures, improving accuracy by 19.2 accuracy points while using \textbf{77.4\% fewer tokens} in the step-wise analysis, as in our pipeline we exchange only selected findings instead of full reasoning traces. 

The key contributions of our work:
\begin{itemize}[leftmargin=*]
	\item \textbf{Shared Memory Buffer:} We propose Muffin, an LLM-based multi-agent workflow in which agents asynchronously exchange selected intermediate findings through a shared buffer.
    \item \textbf{Asynchronously Selective Information Sharing:} Muffin keeps each agent's local reasoning information private and shares partial intermediate findings through the buffer asynchronously, enabling agent collaboration without exposing full reasoning traces.
    \item \textbf{Accuracy-Cost Trade-off:} Muffin outperforms single-agent and majority-vote baselines across four reasoning benchmarks, while lying on the accuracy-token Pareto frontier. 
\end{itemize}

%% file: Content/related_works.tex
\section{Related work}

\paragraph{Multi-agent collaboration.}
Recent research has shown that agentic systems can improve reasoning and performance by organizing multiple specialized agents rather than relying on a single language model inference~\citep{hong2024metagpt,AutoAgentsijcai,llmbasedagentsurvey,Abou_2025_AgenticAISurvey}.
Several systems instantiate this idea through task decomposition~\citep{AgentVerse_ICLR2024,qian-etal-2024-chatdev}, where agents will have specialized roles while working on different parts of a task, including general reasoning workflows and software development pipelines.
General-purpose agent frameworks make this coordination more flexible by allowing LLM-backed agents, tools, and humans to be composed into different interaction patterns~\citep{wu2024autogen,moura2023crewai,langgraph2024}.
Despite these differences, a common denominator of most multi-agent systems is that they still integrate outputs only after full messages, subtasks, or rounds have completed, limiting how agents can adapt to one another's reasoning while inference is still unfinished.

\textbf{Aggregation and shared intermediate information.}
Some studies on aggregation show how outputs from other inference processes can be used as additional context for improving final answers~\citep{jiang-etal-2023-llm,wang-etal-2024-unleashing}.
Layered aggregation uses responses from earlier model calls to condition later ones, enhancing the complement of strengths across models~\citep{Mixture_of_Agents_ICLR2025}.
Debate-based systems instead expose agents to peer answers and critiques over multiple rounds before selecting a final answer~\citep{multiagent_debate_ICML2024,liang-etal-2024-encouraging,xiong-etal-2023-examining}.
These results support the value of shared intermediate information, but the shared object is usually a complete response, critique, or round output, rather than having the LLM judge which information is useful enough to share.
This leaves open how agents should share and use partial information before a full answer has been completed.

\textbf{Fine-grained and streaming interaction.}
More recent work studies collaboration below the level of complete turns through token-level collaboration~\citep{hsu2025groupthinkmultipleconcurrent}, which lets other agents observe generation progress and adjust their reasoning before a complete answer is produced.
Similarly, streaming communication sends reasoning steps to downstream agents during generation, reducing the delay caused by waiting for complete turns~\citep{ICLR2024_SoT,yang2026streamingcommunicationmultiagentreasoning}.
This direction suggests that sharing partial reasoning during inference can reduce coordination latency and enable earlier reactions, while leaving room to study how partial information can be selected, maintained, and jointly reused.
A key challenge is to support asynchronous sharing, where agents can make useful findings available to others without exposing their full reasoning traces. This frames a middle ground between isolated final-answer aggregation and full-trace exchange.

%% file: Content/03_methodology.tex
\section{Methodology}

\subsection{General Multi-Agent Problem Setting}
\label{subsec:problemSetting}
In a multi-agent system, a team of $N$ agents $\mathcal{A}=\{a_i\}_{i=1}^N$ tries to solve a given query $x$. We initialize each agent $a_i$ with a language model $p_{\theta_i}$ having an agent-specific role $r_i$ and a subtask $\tau_i$, executing $T$ reasoning steps. An agent holds a local history $h_i^{(t)}$ for each step $t$, and conditions on a prompt $c_i^{(t)}$ constructed by the shared context formatting function $C_{agent}$, i.e.:
$$
c_i^{(t)} = C_{agent}(x, r_i, \tau_i, h_i^{(t)})
$$
The message generated by the agent is denoted as $m_i^{(t)}$:
\[
m_i^{(t)} \sim p_{\theta_i}(\cdot \mid c_i^{(t)}).
\]
In standard multi-agent inference, agents usually communicate using completed messages.
%
The agents can work sequentially and  respond one after another~\citep{hong2024metagpt}, or in parallel, where multiple agents generate their outputs independently before a final aggregation or selection step~\citep{Mixture_of_Agents_ICLR2025}.

After one or more inference rounds, a final answer is produced by an aggregation function $G$, which may select one agent's output, combine multiple outputs, apply a majority vote, or synthesize a new response:
%
$$
y = G(c_{\mathrm{sum}}) , \qquad \text{with~} c_{\mathrm{sum}} = C_{\mathrm{sum}}(x, m_1^{(1:T)}, \ldots, m_N^{(1:T)})
$$
%
%
where $c_{\mathrm{sum}}$ is the aggregation context constructed from the query and completed agent messages by the formatting function $C_{\mathrm{sum}}$.
Therefore, the standard setting treats communication mainly as completed messages rather than explicitly modeling how information from each round is organized and reused in later inference steps.

\subsection{Muffin - Shared memory buffer Inference}
\label{subsec:muffin}
To support asynchronous sharing of intermediate findings, we introduce a shared memory buffer. First, an orchestrator routes the input query according to the expected demands to a single-agent or multi-agent setting. In the latter case, it configures a team of agents. 
The agents reason concurrently and exchange insights through a shared memory buffer. Finally, a summarizer produces the final answer. 
Our general workflow is illustrated in \Cref{fig:overall_workflow}. 

\textbf{Orchestrator.}
The orchestrator is a language model $p_{\theta_O}$ that maps a query $x$ to an inference plan:
$$
\pi = (z, \{\alpha_i\}_{i=1}^{N}, \rho) \sim p_{\theta_O}(\cdot \mid x), \qquad \text{with~}\alpha_i =(a_i, r_i, \tau_i),
$$
where $z \in \{\mathrm{direct}, \mathrm{multi}\}$ is the routing decision, $\alpha_i$ assigns each agent $a_i$ a role $r_i$ and a non-overlapping subtask $\tau_i$, and $\rho$ denotes the collaboration protocol. The protocol specifies what kind of information is shared (e.g., critique, warnings, etc.), defines what to keep only in an agent's local reasoning sphere, and sets whether new findings may trigger reactive steps of other peers. In case of $z=\mathrm{direct}$, a single call $y \sim p_\theta(\cdot | x)$ terminates the pipeline without memory buffer or summarizer. For cases where more demanding reasoning is expected, the pipeline triggers the multi-agent route.

\textbf{Asynchronous Multi-Agent Reasoning via a Shared Memory Buffer.}
In the multi-agent route, the set of agents can share insights via a shared memory buffer. 

\textbf{\textbullet~Shared Memory Buffer.} Due to an asynchronous sharing of information over a shared memory buffer, we refer to the memory's state at time step $d$ with $\ell$-many entries as:
$$
B^{(d)} = (e_1, \ldots, e_{\ell}), \qquad \text{with~} e_j=(a_i, s_i^{(t)},t) \qquad\text{and~} B^{(0)}=\emptyset.
$$
Each entry $e_j$ stores the information about the source agent $a_i \in \mathcal{A}$, the agent's finding in the $t$-th reasoning step $s_i^{(t)}$, and an identifier of the source step $t$. We define the memory buffer as an append-only data structure, i.e., any existing data is immutable, and the only permitted mutation is adding new information. Hence, $B^{(d)} {\preceq} B^{(d')} $ for two time stamps $d \leq d'$. Since agents can only append information, there are no write conflicts and no locking beyond atomicity of the append operation. When retrieving information from the buffer, agents access the stored data through a bounded window denoted as $R_k(B^{(d)},i)$ to retrieve the $k$ most recent findings of $B$ at time $d$ contributed by other agents. In particular, it scans backward and skips the agent's $a_i$ own shared findings, which are available in the agent's private history.

\begin{figure}[t]
    \centering
    \includegraphics[width=1\linewidth]{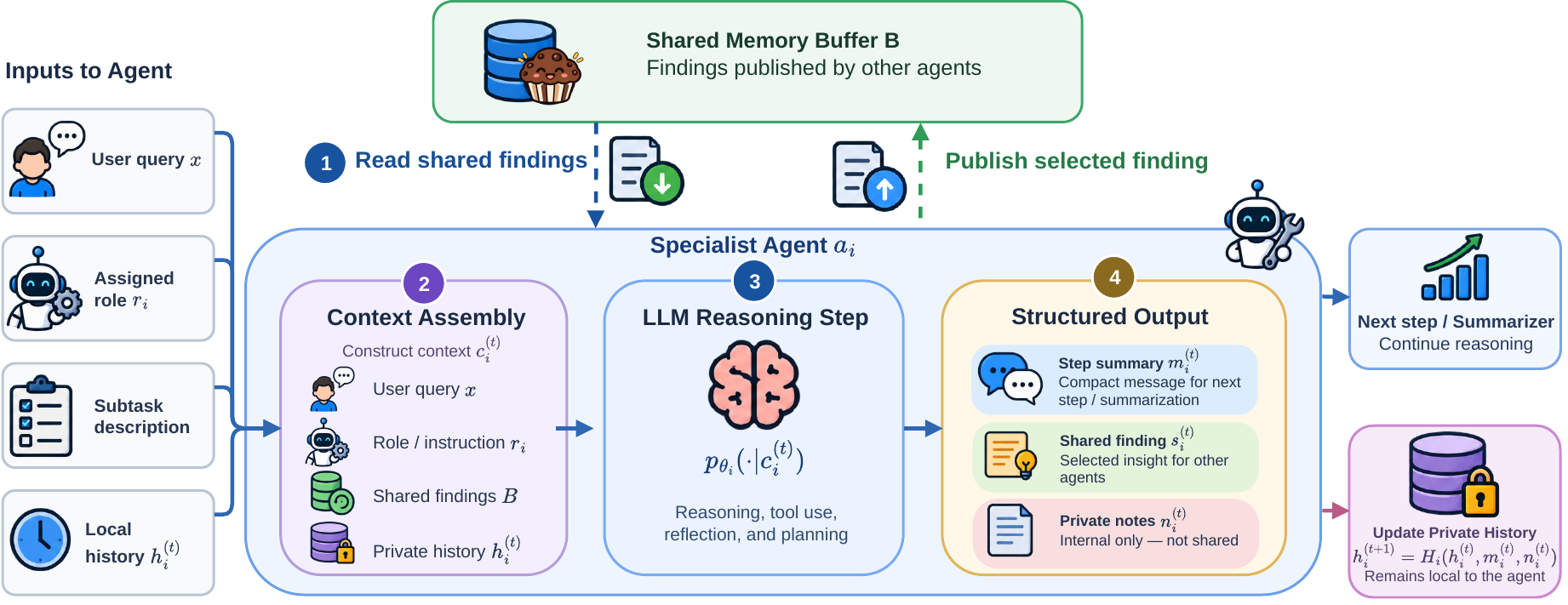}
    \caption{An agent's internal workflow. }
    \label{fig:internal_workflow}
\end{figure}

\textbf{\textbullet~Agent step.} The agent's internal workflow is summarized in \Cref{fig:internal_workflow}. At reasoning step $t$, we denote the agent's dispatch time as $\beta_{i,start}^{(t)}$. Given a generation latency of $\delta_i^{(t)}$, the agent ends the $t$-th inference step at $\beta_{i,end}^{(t)}=\beta_{i,start}^{(t)}+\delta_i^{(t)}$. At dispatch, the agent reads the $k$ most recent findings of other peers from the memory buffer and constructs its context information:
$$
c_i^{(t)} = C_{agent}(x, r_i, \tau_i, h_i^{(t)}, R_k(B^{\beta_{i,start}^{(t)}},i), \rho),
$$
where $h_i^{(t)}$ denotes the agent's local history.
An agent generates a structured output according to:
$$
(m_i^{(t)}, s_i^{(t)}, n_i^{(t)}) \sim p_{\theta_i}(\cdot | c_i^{(t)}),
$$
where $m_i^{(t)}$ denotes an agent's step summary, $s_i^{(t)}$ refers to the optional shared finding, and $n_i^{(t)}$ are local notes only visible for the respective agent. In case of a finding is shared, the buffer is updated:
$$
B^{\beta_{i,end}^{(t)}} = B^{\beta_{i,end}^{(t)-}} \circ (a_i, s_i^{(t)}, t),
$$
where $\circ$ appends a new entry to the memory buffer, $B^{\beta_{i,end}^{(t)-}}$ denotes the buffer state immediately before agent $a_i$ commits its new finding.
The private history is updated independently,
$$
h_i^{(t+1)} = H_i(h_i^{(t)}, m_i^{(t)}, n_i^{(t)})
$$
The history update $H_i$ retains only a compact summary rather than a full trajectory to keep the agents' local states small.

\textbf{\textbullet~Asynchronous execution.} In contrast to prior research, our work does not rely on a global round index. Specifically, an agent $a_i$ is dispatched $\beta_{i,start}^{(t)}$ as soon as it completes the last committing $\beta_{i,end}^{(t-1)}$, independently of the progress of any other agent. Therefore, agents run as concurrent requests and the generation $\delta_i^{(t)}$ dominates latency during which other agents continue to read and write from the buffer. W.l.o.g., assuming two agents $a_i, a_j \in \mathcal{A}$ and two reasoning time steps $t, t'$, a finding that is committed at $\beta_{i,end}^{(t)}$ by the $i$-th agent is visible for other agent $a_j$ iff $\beta_{i,end}^{(t)} \leq \beta_{j,start}^{(t')}$, and lies within the $k$ most recent findings. 
Hence, any two overlapping intervals $[\beta_{i,start}^{(t)}, \beta_{i,end}^{(t)}]$ and $[\beta_{j,start}^{(t')}, \beta_{j,end}^{(t')}]$ with $(\beta_{j,start}^{(t')}>\beta_{i,start}^{(t)})\land(\beta_{j,start}^{(t')} < \beta_{i,end}^{(t)})$ cannot share new findings that are not yet committed. If the collaboration protocol $\rho$ enables re-activation, an agent executes exactly one additional reasoning step and we define the agent's own reasoning horizon as $T_i=T{+}1$ dispatched at $\beta_{i,start}^{(T+1)}=\max_j (\beta_{j,end}^{(T)})$, i.e., when all agents finished at least T reasoning steps. 

\textbf{Summarizer.}
After each agent finishes its reasoning up to $T_i$ steps, our system builds a structured input for the summarizer rather than giving the full interaction trace from each agent. We feed the summarizer the query $x$, the agents' assignments $\alpha_i$, and the last step summary $m_i^{(T_i)}$ of each agent. Since each $m_i^{(T_i)}$ was produced after reading the buffer, findings from the buffer $B$ reach the \textit{Summarizer} module through the agents' summaries. The answer is produced by a separate language model $p_{\theta_{\mathrm{sum}}}$ (in our experiments the same served model):
$$
y \sim p_{\theta_{\mathrm{sum}}}\!\left( \cdot \mid c_{\mathrm{sum}} \right),
\qquad \text{with } c_{\mathrm{sum}} = C_{\mathrm{sum}}\!\left( x, \left\{ (\alpha_i, m_i^{(T_i)}) \right\}_{i=1}^{N} \right)
$$
The summarizer is focusing on the task of choosing and summarizing the best-supported answer from the agents' last message summaries, using the candidate summary as evidence rather than trusting it completely.
It is also instructed not to solve the task again, introduce new assumptions, or choose an answer which is not supported by the agent outputs unless no candidate can be extracted.
We analyze the summary's design choice in Section~\ref{sec:ablations} by comparing the summarizer output with the agents' last individual outputs.

%% file: Content/experiments.tex
\section{Experiments}
\label{sec:experiments}

Our experiments are organized around three questions. First, we evaluate whether our method improves reasoning accuracy.
Second, we study the inference-cost trade-off introduced by multi-agent collaboration. 
Third, we analyze whether the shared buffer helps agents use intermediate findings from other agents before final summarization.

\subsection{Setup}
\label{subsec:setup}
We first describe the overall experimental setting, including the benchmarks, baselines, and implementation details.

\textbf{Benchmarks.}
We evaluate on four benchmarks that cover complementary reasoning tasks. The first two follow the evaluation protocol of prior work~\citep{multiagent_debate_ICML2024}:
\begin{itemize}[leftmargin=*]
    \item \textbf{Chess Move Validity}, derived from the BIG-Bench Chess State Tracking benchmark~\citep{srivastava2023beyond}, measures a system's predictive performance of a valid move in a chess setting.
    \item \textbf{MMLU-Pro}~\citep{mmlupro} for broad multi-domain multiple-choice reasoning.
    \item \textbf{GPQA-Diamond}~\citep{rein2024gpqa} tests difficult, knowledge-intensive scientific reasoning.
    \item \textbf{OlymMATH}~\citep{sun2025challengingboundariesreasoningolympiadlevel}, which evaluates mathematical problem solving that often requires multi-step derivation, proof-like reasoning, and verification.
\end{itemize}
For Chess Move Validity and MMLU-Pro, we use the 100 examples from each dataset, following the 100-example evaluation setting of prior work~\citep{multiagent_debate_ICML2024}. 
For GPQA-Diamond, we evaluate on the full Diamond split of 198 examples.
For OlymMATH, we use 100 examples from the English easy (-E) subset and 100 examples from the English hard (-H) subset, and report the results for the two subsets separately.

\textbf{Baselines.}
We compare Muffin with single-agent, majority-voting based multi-agent, and multi-agent debate:
\begin{itemize}[leftmargin=*]
    \item \textbf{Single-Agent} uses only one language model agent to solve the task. We explicitly allow for multiple reasoning steps and not just one-shot inference.
    \item \textbf{Majority Vote} leverages multiple independent single-agent, and gathers their final results and choose the majority answer from them.
    \item \textbf{Multi-Agent Debate} follows the setting of the Multi-agent Debate~\citep{multiagent_debate_ICML2024},
    where the agents will revise their responses after observing other agents' output.
\end{itemize}

\textbf{Implementation Details.}
We evaluate all methods with four LLMs:
\texttt{gemma-4-E4B-it}, \texttt{gemma-4-31B-it},
\texttt{Qwen3.8-27B}, and \texttt{Qwen3.5-4B}~\citep{gemmateam2026gemma4,qwen38,qwen3.5}.
All models are served on a compute cluster through a \texttt{vLLM}~\citep{kwon2023efficient} OpenAI-compatible endpoint. All baselines and our method call the same served model in each comparison.
We serve the models using tensor parallelism from \texttt{vLLM}, utilizing two H100 GPUs for the smaller models and four H100 GPUs for the larger models. As LLM inference might introduce small run-to-run difference, we repeat each experiment three times and report averaged results. For method execution and agent parameters, unless explicitly stated otherwise, we set the number of steps to five for all methods and fix the number of agents to three for Majority Vote, Multi-Agent Debate, and Muffin.
Our code is publicly available: \url{https://anonymous.4open.science/r/Muffin}

\subsection{Effectiveness and Efficiency}

\textbf{Hypothesis 1: Asynchronous sharing of intermediate findings through a memory buffer improves accuracy over single-agent, majority-vote, and multi-agent debate baselines under the same step budget. 
}\hfill\\
For the first experiment, we want to support the hypothesis that sharing intermediate findings rather than the full output from others can improve accuracy. Under the default setting defined in Section~\ref{subsec:setup}, we compare the task performance of Muffin against the baselines across all selected benchmarks and models.

\begin{table}[t]
\centering
\caption{Accuracy by model, method, and benchmark. For OlymMATH, we report the English easy (OlymMATH-E) and English hard (OlymMATH-H) subsets separately. For each model and benchmark, the best result is in \textbf{bold} and the second-best is \underline{underlined}. The value after $\pm$ denotes the standard deviation across three independent runs. $^{\dagger}$Low due to the answer-extraction failures, counted as incorrect.}
\label{tab:hypothesis1_accuracy}
\small
\setlength{\tabcolsep}{4pt}
\resizebox{\textwidth}{!}{
\begin{tabular}{clccccc}
\hline
\textbf{Model} & \textbf{Method} & \textbf{Chess} & \textbf{GPQA} & \textbf{MMLU-Pro} & \textbf{OlymMATH-E} & \textbf{OlymMATH-H} \\
\hline
gemma-4-E4B-it & Single-Agent & \accstd{0.223}{0.006} & \accstd{0.513}{0.007} & \accstd{0.740}{0.020} & \accstd{0.350}{0.000} & \accstd{0.073}{0.006} \\
 & Majority Vote & \accstd{0.260}{0.000} & \accstd{0.534}{0.003} & \accstd{0.750}{0.010} & \accstd{0.350}{0.000} & \accstd{0.073}{0.006} \\
 & Multi-Agent Debate & \accstdunderline{0.300}{0.000} & \accstdbf{0.658}{0.010} & \accstdunderline{0.770}{0.000} & \accstdunderline{0.423}{0.003} & \accstdunderline{0.103}{0.006} \\
 & Muffin & \accstdbf{0.310}{0.013} & \accstdunderline{0.632}{0.030} & \accstdbf{0.790}{0.020} & \accstdbf{0.593}{0.012} & \accstdbf{0.133}{0.032} \\
\hline
gemma-4-31B-it & Single-Agent & \accstd{0.660}{0.000} & \accstd{0.691}{0.054} & \accstdunderline{0.850}{0.000} & \accstd{0.703}{0.021} & \accstd{0.190}{0.010} \\
 & Majority Vote & \accstd{0.660}{0.000} & \accstdunderline{0.715}{0.043} & \accstdunderline{0.850}{0.000} & \accstd{0.707}{0.012} & \accstd{0.193}{0.015} \\
 & Multi-Agent Debate &\accstdbf{0.850}{0.000} & \accstd{0.379}{0.016}$^{\dagger}$ & \accstdunderline{0.850}{0.000} & \accstdunderline{0.767}{0.040} & \accstdunderline{0.383}{0.021} \\
 & Muffin & \accstdunderline{0.793}{0.021} & \accstdbf{0.828}{0.050} & \accstdbf{0.863}{0.006} & \accstdbf{0.850}{0.010} & \accstdbf{0.407}{0.006} \\
\hline
Qwen3.5-4B & Single-Agent & \accstd{0.283}{0.006} & \accstd{0.599}{0.028} & \accstd{0.720}{0.095} & \accstd{0.370}{0.020} & \accstd{0.120}{0.035} \\
 & Majority Vote & \accstd{0.380}{0.010} & \accstdunderline{0.655}{0.031} & \accstdunderline{0.757}{0.076} & \accstdunderline{0.397}{0.006} & \accstd{0.130}{0.035} \\
 & Multi-Agent Debate & \accstdunderline{0.410}{0.010} & \accstd{0.567}{0.025} & \accstd{0.740}{0.095} & \accstdunderline{0.397}{0.006} & \accstdunderline{0.213}{0.015}\\
 & Muffin & \accstdbf{0.560}{0.020} & \accstdbf{0.712}{0.040} & \accstdbf{0.780}{0.062} & \accstdbf{0.597}{0.045} & \accstdbf{0.237}{0.021} \\
\hline
Qwen3.8-27B & Single-Agent & \accstd{0.987}{0.015} & \accstd{0.859}{0.010} & \accstd{0.840}{0.020} & \accstd{0.790}{0.010} & \accstd{0.670}{0.010} \\
 & Majority Vote & \accstdunderline{0.993}{0.006} & \accstdbf{0.897}{0.020} & \accstdunderline{0.853}{0.031} & \accstd{0.797}{0.015} & \accstd{0.680}{0.010} \\
 & Multi-Agent Debate & \accstd{0.990}{0.010} & \accstd{0.879}{0.016} & \accstd{0.817}{0.049} & \accstdunderline{0.830}{0.030} & \accstdunderline{0.683}{0.015} \\
 & Muffin & \accstdbf{1.000}{0.000} & \accstdunderline{0.891}{0.030} & \accstdbf{0.860}{0.036} & \accstdbf{0.850}{0.035} & \accstdbf{0.710}{0.026} \\
\hline
\end{tabular}
}
\end{table}

Table~\ref{tab:hypothesis1_accuracy} shows that Muffin achieves the best accuracy in 17 out of 20 in our evaluation settings and is second-best in the remaining three. The largest improvements appear on the OlymMATH benchmark, and relative to the strongest baseline, Muffin improves accuracy by 11.8 points on OlymMATH-E and 2.6 points on OlymMATH-H on average. On benchmarks where the baselines perform better, the margin is smaller, showing that the main benefit appears when the task leaves room for multi-step collaborative reasoning.
Muffin also shows slightly higher standard deviations in some settings, which is expected because dynamic agent roles and LLM-decided shared-finding exchange introduce more variation across runs.

\textbf{Hypothesis 2: Asynchronous problem solving provides a better quality-cost trade-off.}\hfill\\
We evaluate this hypothesis through three complementary comparisons on the OlymMATH-Easy, a representative benchmark that requires multi-step mathematical reasoning and allows us to compare how accuracy and token usage evolve across inference steps. First we compare different inference methods over ten reasoning steps. Second, we vary the number of agents in Muffin to check whether adding more agents improves accuracy enough to cover the additional token cost.
Third, we compare Muffin with full-trace sharing to evaluate the benefit of selective sharing. In terms of token usage, we measure both input and output tokens. Regarding accuracy, baseline results can be obtained directly after each step, whereas in our method, we extract the outputs at each step and take a majority vote across them.
For both accuracy and token usage, we report the average over three independent runs.

\begin{figure}[t]
    \begin{minipage}[t]{0.49\textwidth}
        \centering
        \includegraphics[width=1\linewidth]{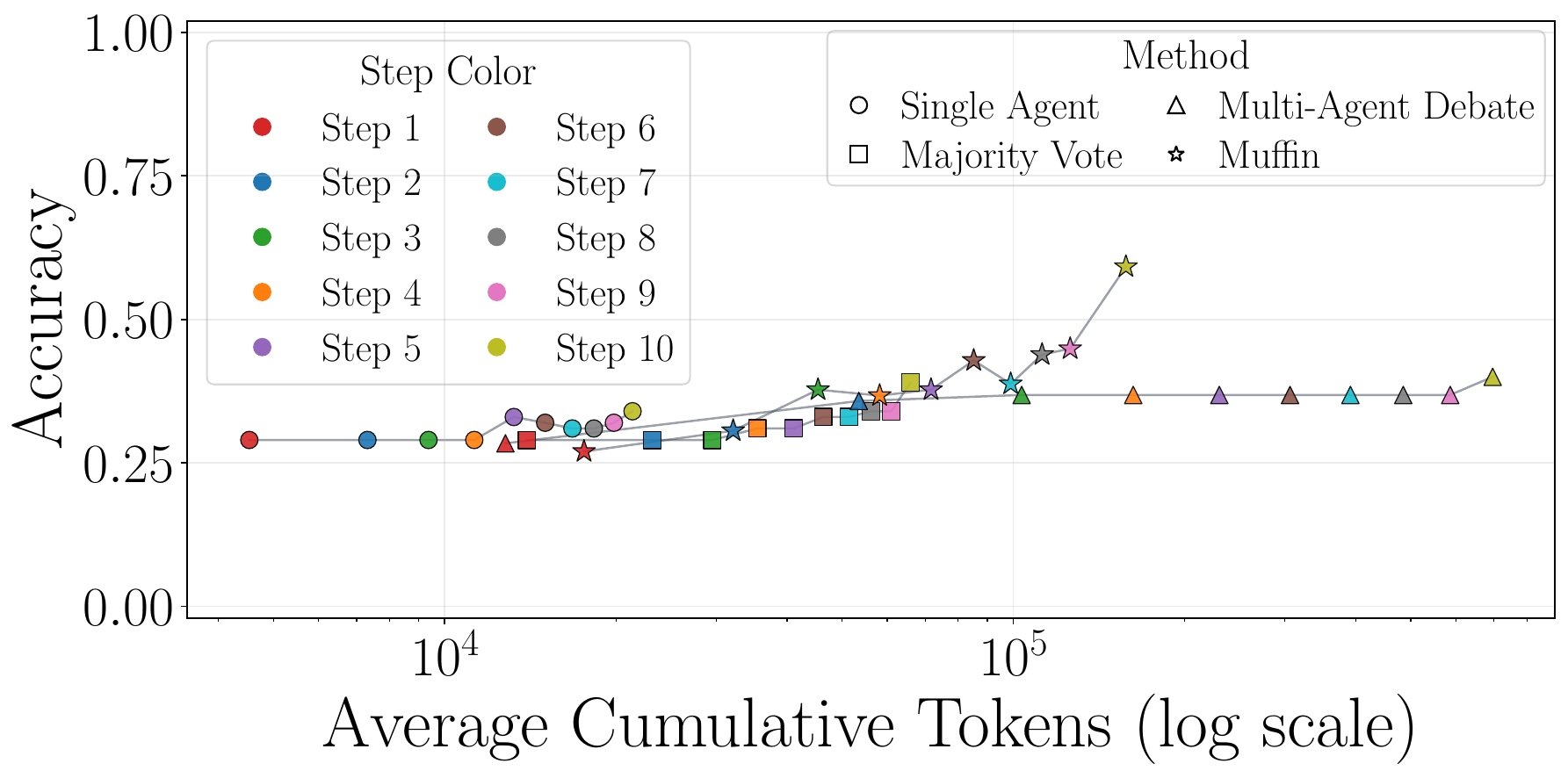}
        \caption{Step-wise accuracy and token usage on OlymMATH-Easy with \texttt{gemma-4-E4B-it}.}
        \label{fig:olymmath_easy_step_token_accuracy}
    \end{minipage}
    \hfill
    \begin{minipage}[t]{0.49\textwidth}
        \centering
        \includegraphics[width=1\linewidth]{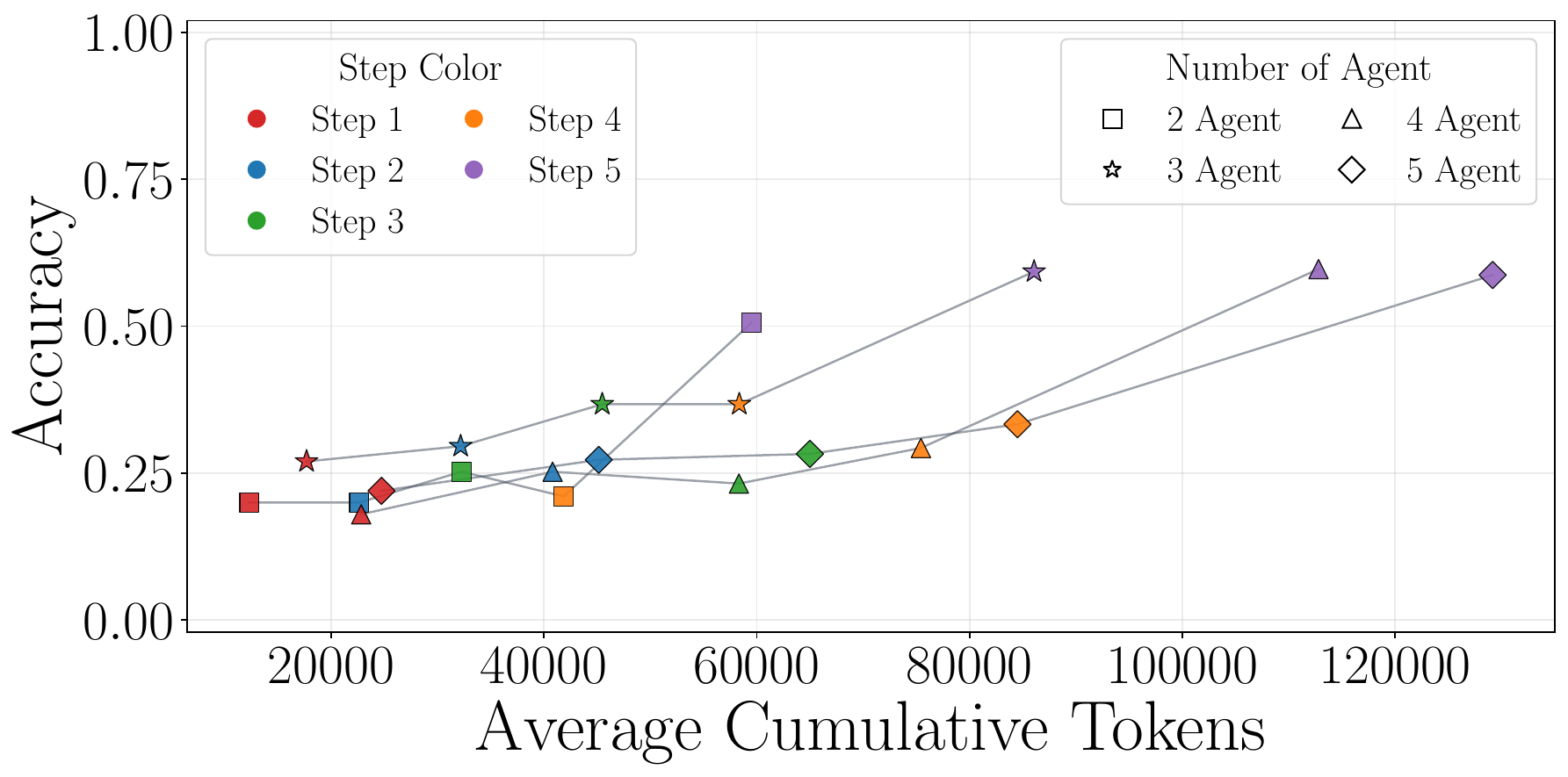}
        \caption{Accuracy-token trade-off on OlymMATH-Easy when varying the number of agents with \texttt{gemma-4-E4B-it}.}
        \label{fig:olymmath_easy_number_of_agent_accuracy}
        \end{minipage}
\end{figure}

\textbullet~\textbf{Method comparison over inference rounds.}
As shown in Figure~\ref{fig:olymmath_easy_step_token_accuracy}, sharing the intermediate findings provides a better quality-cost trade-off on the selected benchmark.
Each symbol represents a different method and reasoning steps are colorized, respectively.
All methods improve in accuracy as the number of reasoning steps increases, with our approach achieving the highest performance by the $6$-th step and gaining a boost in the final round after incorporating the summarizer, increasing from 44.9\% to 59.2\%.
Overall, Muffin reaches the highest final accuracy and dominates the multi-agent baselines, outperforming the strongest baseline by 19.2 accuracy points while using 77.4\% fewer tokens than multi-agent debate.
The remaining benchmarks follow the same trend, with detailed results reported in Appendix~\ref{app:step_token_acc}.

\textbullet~\textbf{Effect of the number of agents.}
We next study the effect of the number of agents on the performance of Muffin. We vary the number of agents from two to five while fixing the reasoning to five steps.
The results shown in Figure~\ref{fig:olymmath_easy_number_of_agent_accuracy} demonstrate that moving from two to three agents improves accuracy from 50.6\% to 59.3\%, showing a clear benefit from additional collaboration.
However, increasing to four or five agents gives only slight performance changes, reaching 59.7\% and 58.7\%, but token usage continues to increase.
This indicates that the benefit of the collaboration does not scale linearly with the number of agents and shows a convergence behavior where additional token usage does not pay off.

\textbullet~\textbf{Full-trace sharing versus our method.}
Lastly, we compare our method with a more standard full-trace approach. As described in \Cref{subsec:muffin}, Muffin only passes local notes but not the full reasoning content to the next step. The \textit{Summarizer} only receives the last output from each agent but not their full reasoning trace. 
In this experiment, we disable the selective sharing and pass the full trace to both the agents and the summarizer to evaluate whether the additional context improves the quality-cost trade-off.

\begin{figure}[t]
    \begin{minipage}[t]{0.49\textwidth}
        \centering
        \includegraphics[width=1\linewidth]{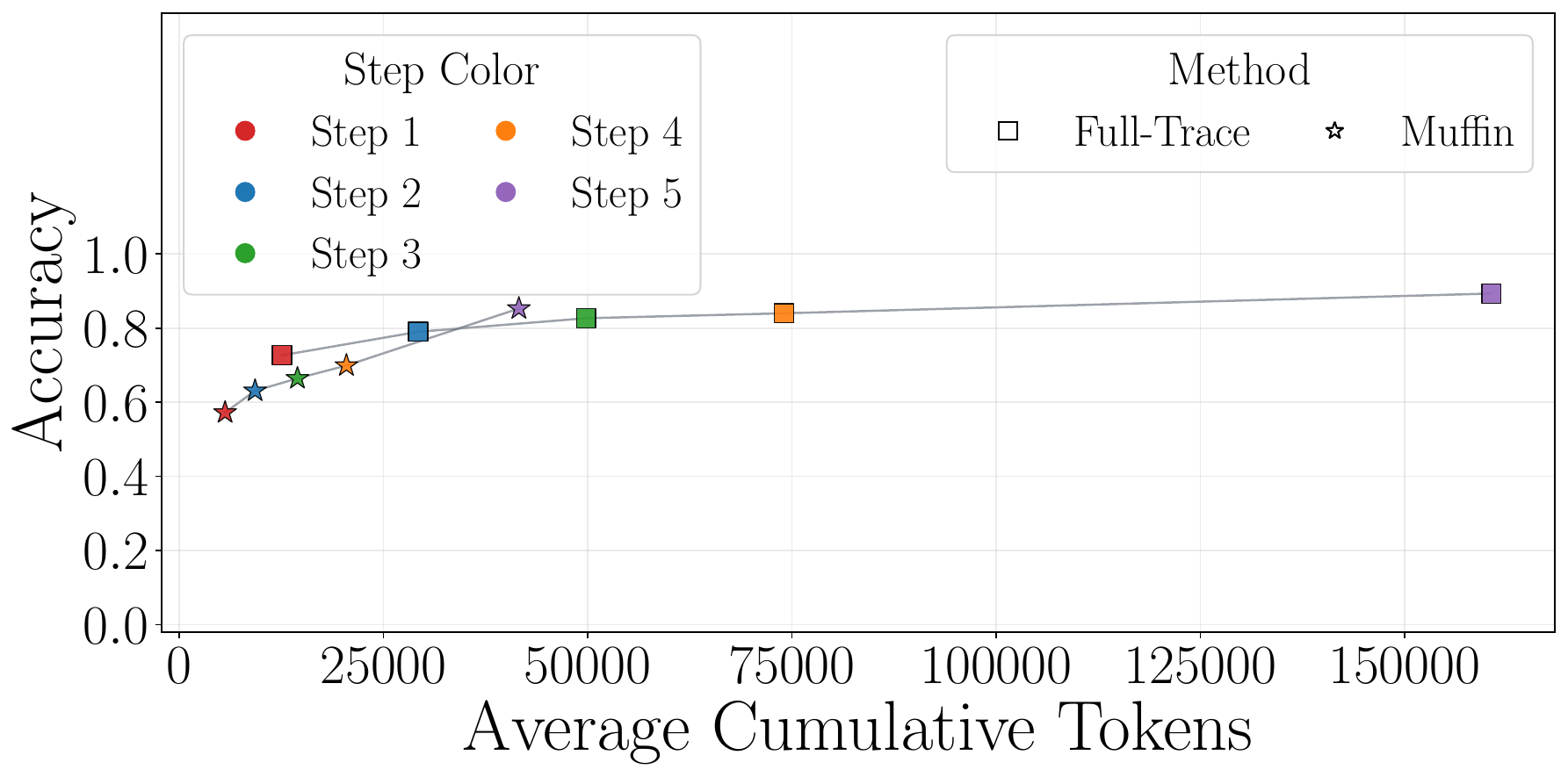}
        \caption{Accuracy-token trade-off on OlymMATH-Easy on full-trace sharing with \texttt{gemma-4-31B-it}.}
        \label{fig:full_trace_vs_our_method}
    \end{minipage}
    \hfill
    \begin{minipage}[t]{0.49\textwidth}
        \centering
        \includegraphics[width=1\linewidth]{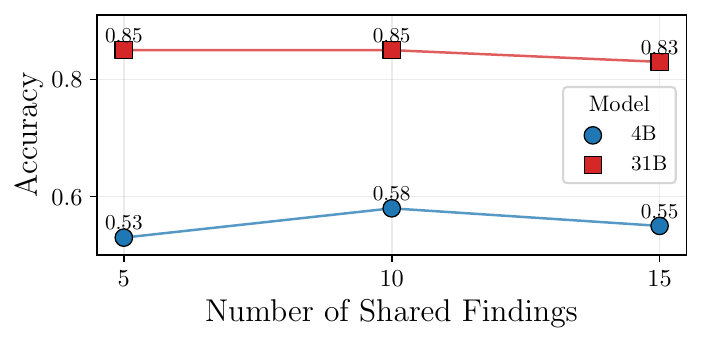}
        \caption{Accuracy under different shared-finding budgets with both sizes of \texttt{gemma-4}.}
        \label{fig:number_of_shared_findings}
        \end{minipage}
\end{figure}

Figure~\ref{fig:full_trace_vs_our_method} shows that the full-trace approach reaches higher accuracy in the early steps, but it requires substantially more tokens.
At the final step, full-trace sharing improves accuracy from 85.0\% to 87.3\%, while Muffin reduces token usage from 160.6K to 41.9K, corresponding to 73.9\% fewer tokens.
This suggests that passing the full reasoning trace can add some useful information, but much of this benefit can be recovered by sharing compact intermediate findings.
Overall, Muffin provides a better quality-cost trade-off than the full-trace variant.

\begin{figure}[t]
  \begin{minipage}[t]{0.49\textwidth}
    \centering
    \includegraphics[width=\linewidth]{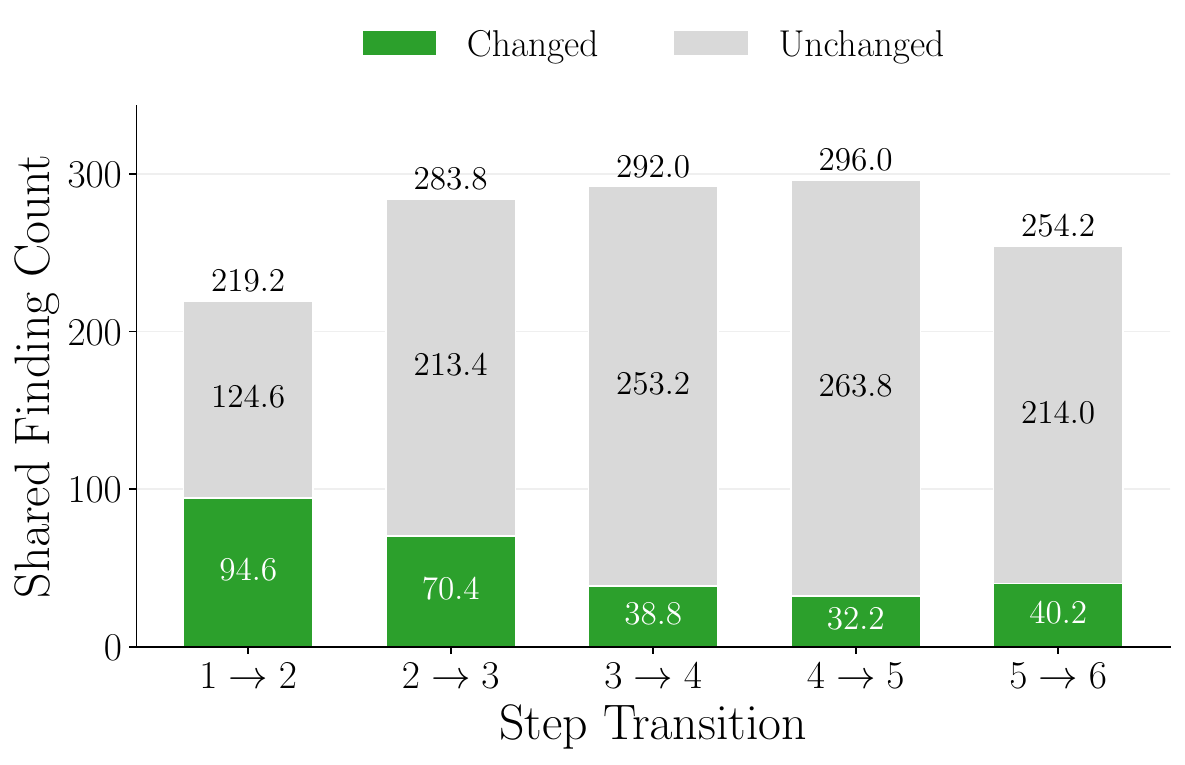}
    \caption{Mean number of shared findings that change or do not change an agent's answer on Chess with \texttt{gemma-4-31B-it}.}
    \label{fig:change_chess}
  \end{minipage}
  \hfill
  \begin{minipage}[t]{0.49\textwidth}
    \centering
    \includegraphics[width=\linewidth]{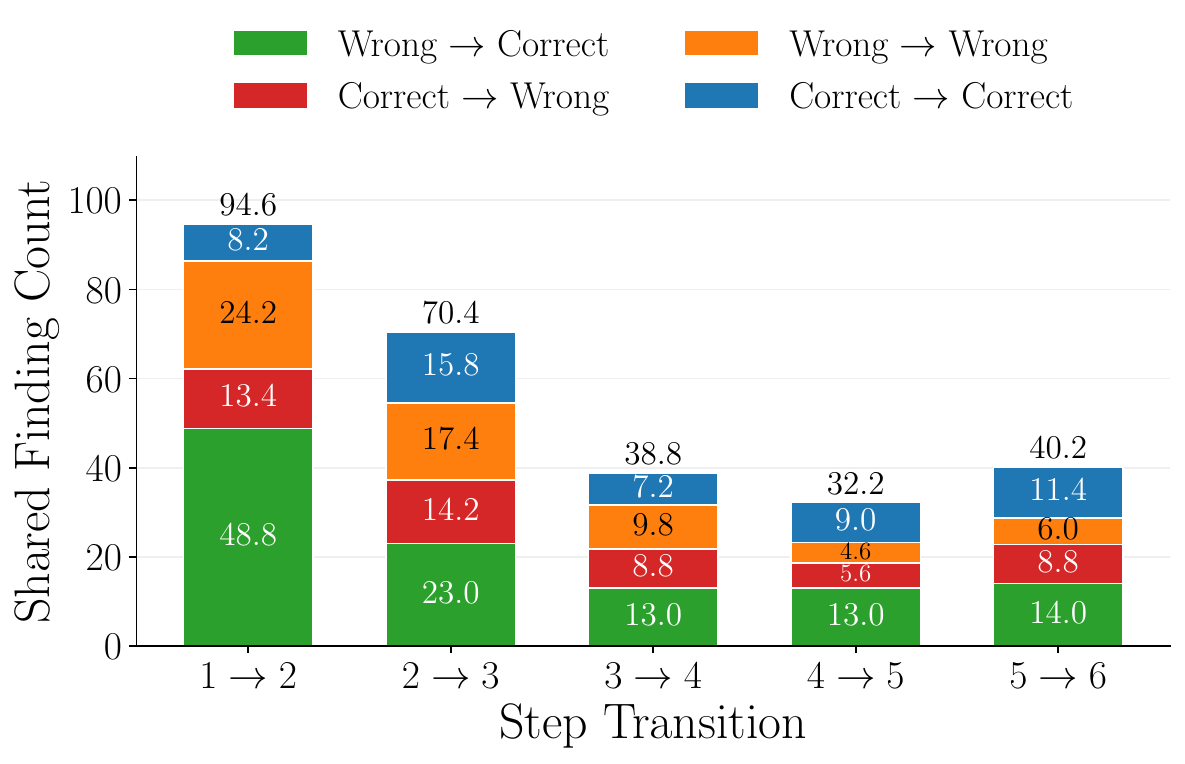}
    \caption{Outcome breakdown for shared findings that change an agent's answer on Chess with \texttt{gemma-4-31B-it}.}
    \label{fig:change_chess_detailed}
  \end{minipage}
\end{figure}

\textbf{Hypothesis 3: Shared findings actively steer the agents' reasoning and move them toward correct answers.}\hfill\\
After each step, we check whether the agents' answers change after a finding is published. Further, we inspect the change in terms of switching from wrong to correct ones, correct to wrong, or whether the correctness status is kept. We report the mean count per run for each step transition.\\
\Cref{fig:change_chess} shows that, on Chess with \texttt{gemma-4-31B-it}, 20.5\% of all findings induced a change in the answer. 
Notably, this insight is highest directly in the first step transition with 43.2\% and decreases over all step transitions to 15.8\% in the last recorded step, indicating agents' convergence as the memory buffer is filled with insights. \\
\Cref{fig:change_chess_detailed} provides more in-depth insights into the changes. Across all step transitions, we observe a majority for wrong-to-correct changes in contrast to correct-to-wrong changes showing a decreasing behavior. The effect is most prominent in the early stage, where 51.6\% turn a wrong answer into a correct one, while only 14.2\% show the opposite. 
In later steps, the shared findings show a more balanced view with 34.8\% vs. 21.9\%. Therefore, early findings shared via the memory buffer have the highest marginal value. The same analysis for the remaining benchmarks with \texttt{gemma-4-31B-it} and \texttt{Qwen3.8-27B} is reported in Appendix~\ref{app:shared_findings}.

\subsection{Ablation Studies}
\label{sec:ablations}
We conduct ablation studies to inspect which parts of our method contribute to the final performance. These experiments focus on two design choices: whether the final summarizer improves over the agents' last inference outputs, and how the number of shared findings affects the performance.

\textbullet~\textbf{Agents' final outputs versus summarizer output.}
We compare the final answer produced by the summarizer with the last individual outputs generated by the agents, which include the last findings from the shared buffer in their input. This evaluates whether the summarizer is giving additional benefit by combining agents' final evidence. We compute this comparison across all benchmarks and two model sizes from the Gemma-4 family. As shown in Table~\ref{tab:last_agent_vs_summarizer}, the summarizer changes the last agent answer in 11.14\% of cases. Among those changed cases, wrong-to-correct changes account for 32.05\%, while correct-to-wrong changes account for 21.20\%.
This means beneficial corrections are 10.85 percentage points more than harmful changes. This suggests that the summarizer does more than copy or choose the majority from the final agent outputs, and can recover correct answers from the agents' final evidence while introducing less harmful noise.

\begin{table}[t]
\centering
\caption{Comparison between the last agent output and the final summarizer output.}
\label{tab:last_agent_vs_summarizer}
\small
\begin{tabular}{lc|lc}
\hline
\multicolumn{2}{c|}{\textbf{All examples}} &
\multicolumn{2}{|c}{\textbf{Changed examples only}} \\
\hline
\textbf{Category} & \textbf{\%} & \textbf{Category} & \textbf{\%} \\
\hline
Same as last agent & 83.62 &
Wrong $\rightarrow$ Correct & 32.05 \\
Different from last agent & 11.14 &
Correct $\rightarrow$ Wrong & 21.20 \\
Direct route & 5.24 &
Wrong $\rightarrow$ Wrong & 39.26 \\
 & &
Correct $\rightarrow$ Correct & 7.49 \\
\hline
\end{tabular}
\end{table}

\textbullet~\textbf{Number of shared findings.}
Next, we control how much information an agent can retrieve from the buffer, and test whether more shared context always improves collaboration. 
We use OlymMATH-Easy with the Gemma-4 models.
Figure~\ref{fig:number_of_shared_findings} shows that increasing the number of shared findings does not always improve accuracy. For \texttt{gemma-4-E4B-it}, the best result appears at $k=10$, improving over $k=5$ by 5.0 accuracy points, but adding more findings reduces performance. By contrast, \texttt{gemma-4-31B-it} is less sensitive to this budget, with accuracy changing by only 2.0 points. 
This suggests that a moderate number of shared findings is sufficient, while adding more findings can introduce noise by including redundant or outdated findings.

%% file: Content/discussion.tex
\subsection{Limitations and Discussion}
Our results show that shared intermediate findings in an asynchronous manner can improve multi-agent reasoning without agents exchanging or using the full reasoning traces. Our evaluation is focused on question-answering and reasoning benchmarks. 
Future work should evaluate shared buffer collaboration on tasks that require tool use, planning, environment interaction, or long-horizon execution. Our experiments are also limited by compute cost. Although we test multiple models, benchmarks, and agent configurations, we cannot cover all closed- and open-weight models with different experimental settings. Broader evaluation would help to clarify how the shared buffer can adapt across different agentic usages.

%% file: Content/conclusion.tex
\section{Conclusion}
Existing multi-agent debate methods rely on synchronous turn-taking and sharing information across different agents. We introduce \textbf{Muffin}, an agentic \textbf{m}emory b\textbf{uff}er \textbf{in}ference framework in which an orchestrator first routes queries by expected reasoning demand. For more demanding queries, the collaborating agents exchange intermediate findings asynchronously through a shared memory buffer while keeping private reasoning local. Finally, a summarizer produces the final answer from compact agent summaries.
Across four reasoning benchmarks and four language models, Muffin achieves the best accuracy in 17 out of 20 benchmark settings and reaches a better accuracy-token trade-off than multi-agent debate.
Muffin improves final accuracy by 19.2 accuracy points while using 77.4\% fewer tokens than multi-agent debate, showing that selective sharing of findings is a lightweight alternative to full-trace exchange.
Future work includes a dynamic scaling of the number of agents, and extending the buffer to tool-using and long-horizon tasks.

%% file: Content/appendix.tex
\section{Prompt Templates}
\label{app:prompts}

This appendix lists the main prompt templates used by Muffin.

\subsection{Orchestrator Prompt}
\label{app:orchestrator_prompt}

The orchestrator decides whether a query should be answered directly or routed to a multi-agent workflow.
In the multi-agent route, it also assigns agent roles, subtasks, and collaboration rules.

\begin{promptbox}
You are the dynamic subagent orchestrator for a local multi-agent system.
Only valid JSON.

Routing:
- Default to multi_agent mode.
- Use direct mode only when the answer is truthfully 100
- Use multi_agent for uncertainty, ambiguity, missing context, calculation risk, domain reasoning, verification, critique, or multiple viewpoints.
- The reason must explain why direct or multi_agent was selected.
- For direct, name the exact evidence or deterministic property and explain why multi-agent verification would not help.

If multi_agent, create {min_dynamic_subagents} to {max_dynamic_subagents} subagents.
- Do not create overlapping roles or duplicate subtasks.
- Each role must own a distinct evidence source, check, or perspective.
- At least one subagent must challenge assumptions and contradictions.
- Rules should be short, specific operating or communication rules.

For multi_agent, output:
{
  "mode": "multi_agent",
  "task_type": "{short task type}",
  "task_summary": "{one sentence summary}",
  "reason": "{why multi-agent reasoning is needed}",
  "selected_agents": [
    {
      "name": "{agent name}",
      "role": "{agent role}",
      "description": "{unique evidence, constraints, or checks}",
      "rules": ["{operating rule}"],
      "subtask": "{non-overlapping subtask}",
      "expected_output": "{expected output}",
      "critical_debate": false,
      "workspace_access": "none"
    }
  ],
  "collaboration_protocol": {
    "event_types_to_share": ["finding", "critique", "warning"],
    "reactive_steps": true,
    "notes": "{how agents should use shared findings}"
  }
}

For direct, output:
{
  "mode": "direct",
  "task_type": "{short task type}",
  "task_summary": "{one sentence summary}",
  "reason": "{why the answer is certain and verification would not help}",
  "direct_certainty": "100_percent",
  "selected_agents": [],
  "collaboration_protocol": {
    "event_types_to_share": ["finding", "critique", "warning"],
    "reactive_steps": false,
    "notes": "No agent collaboration is needed."
  }
}

User task:
{task}
\end{promptbox}

\subsection{Agent Reasoning Prompt}
\label{app:agent_prompt}

Each agent receives the original task, its role and subtask assignment, its private local notes, and the current shared findings.
The agent can only writes a concise shared finding to the buffer while keeping private reasoning local.

\begin{promptbox}
{task}

Shared protocol:
- Work independently on the assigned subtask.
- Share concise findings that may help other agents.
- Check shared findings before answering.
- Shared findings may include an answer candidate plus a short reason.
- Verify shared findings before adopting them.
- If a shared finding is used or rejected, mention the source briefly in SUMMARY or LOCAL_NOTES.
- Do not reveal private chain-of-thought.
- If assigned as a critical agent, challenge assumptions, contradictions, and overconfident claims.

Output format:
Respond with concise summaries only. Use this exact format:

SUMMARY:
{brief reasoning summary}

SHARE_FINDING:
{one useful finding for other agents, or blank}

LOCAL_NOTES:
ANSWER_CHOICE: {current answer candidate, or "undecided"}
ANSWER_REASON: {one sentence reason for the current answer}

Agent name:
{agent_name}

Agent role:
{role}

Agent description:
{description}

Agent rules:
- {rule_1}
- {rule_2}

Assigned subtask:
{assigned_subtask}

Expected output:
{expected_output}

Current step:
{step_index} of {max_steps}

Local notes:
{notes}

Shared findings:
{shared findings}
\end{promptbox}

\subsection{Summarizer Prompt}
\label{app:summarizer_prompt}

After the multi-agent reasoning steps, the summarizer selects the final answer from the agents' last summaries and extracted answer candidates.

\begin{promptbox}
You are the Summarizer for a dynamic multi-agent benchmark run.

Your job is to choose and summarize the best-supported answer from the agents' last summaries.
Do not solve the task again.
Do not introduce new calculations, new assumptions, or a new answer that was not supported by the agent outputs.

Original task:
{task}

Agent last summaries:
{agents' rule and agents' last summaries}

Candidate aggregation:
{candidate answer aggregation}

Instructions:
- Identify the final answer candidates already present in the agent summaries.
- Use the candidate aggregation as evidence, not as automatic authority.
- If needs_review is false and a selected_candidate exists, keep that candidate unless the agent summaries clearly contradict it.
- If needs_review is true, audit the candidate reasons and choose the best-supported candidate from the agent outputs.
- Treat weak majority, ties, and single-candidate-only cases as requiring evidence review.
- Use each agent's role, subtask, description, and explanation excerpt when candidates conflict.
- Do not introduce a new answer absent from the agent candidate list unless no candidate was extractable.
- If the agent outputs include a clear final-answer format, keep that format.
- Keep the response concise.
- End with exactly one final answer line.
\end{promptbox}

\subsection{Direct-Answer Prompt}
\label{app:direct_prompt}

When the orchestrator selects the direct route, the system bypasses the multi-agent workflow and asks the model to answer the task directly.

\begin{promptbox}
Answer the user task directly with one concise response.

User task:
{task}
\end{promptbox}

\section{Additional Experimental Results}
\label{app:additional_results}

\subsection{Step-wise Accuracy and Token Usage}
\label{app:step_token_acc}

\begin{figure}[H]
    \begin{minipage}[t]{0.49\textwidth}
        \centering
        \includegraphics[width=1\linewidth]{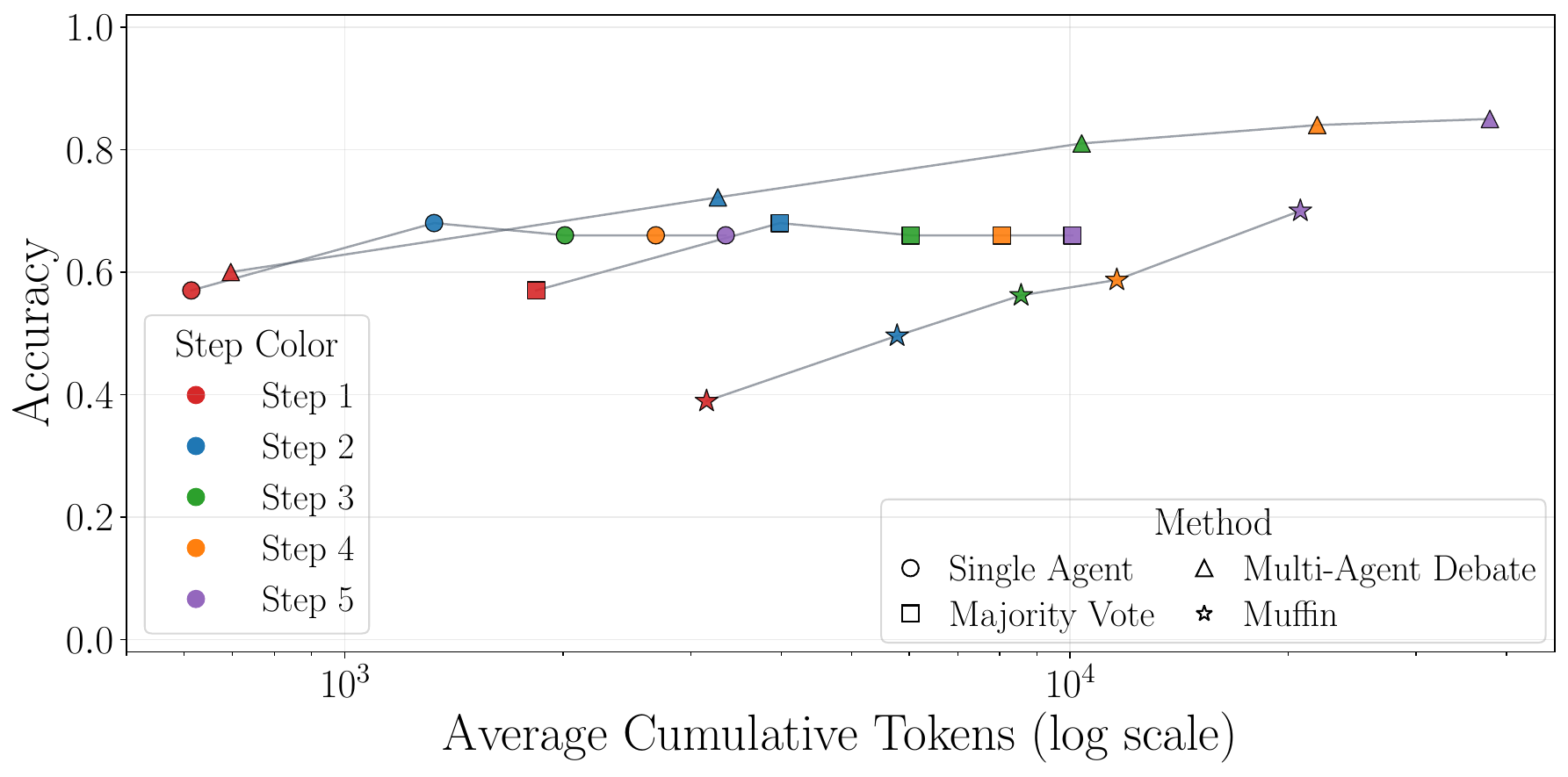}
        \caption{Step-wise accuracy and token usage on Chess with \texttt{gemma-4-31B-it}.}
        \label{fig:app_step_token_chess}
    \end{minipage}
    \hfill
    \begin{minipage}[t]{0.49\textwidth}
        \centering
        \includegraphics[width=1\linewidth]{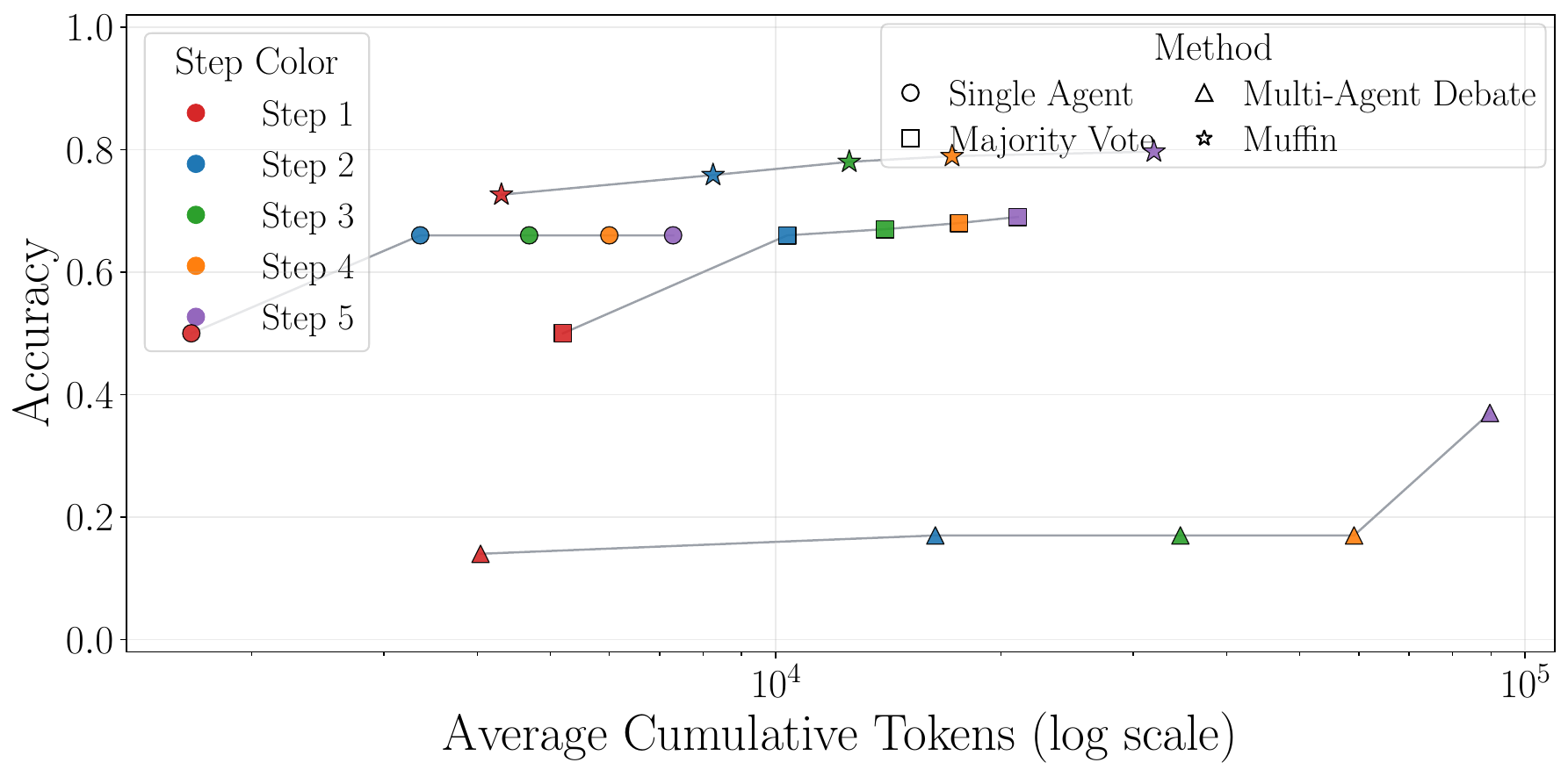}
        \caption{Step-wise accuracy and token usage on GPQA with \texttt{gemma-4-31B-it}.}
        \label{fig:app_step_token_gpqa}
    \end{minipage}
\end{figure}

\begin{figure}[H]
    \begin{minipage}[t]{0.49\textwidth}
        \centering
        \includegraphics[width=1\linewidth]{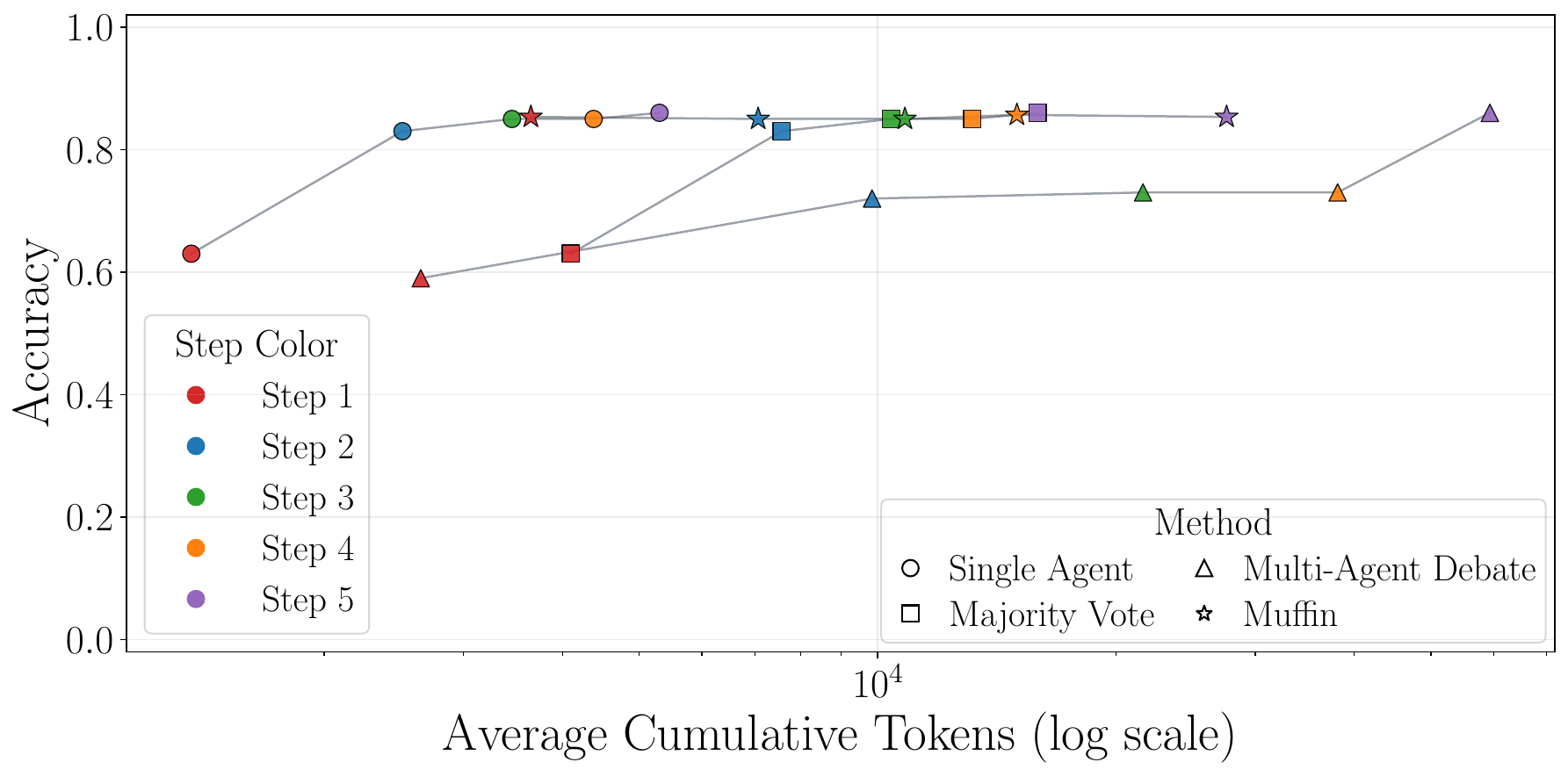}
        \caption{Step-wise accuracy and token usage on MMLU-Pro with \texttt{gemma-4-31B-it}.}
        \label{fig:app_step_token_mmlu_pro}
    \end{minipage}
    \hfill
    \begin{minipage}[t]{0.49\textwidth}
        \centering
        \includegraphics[width=1\linewidth]{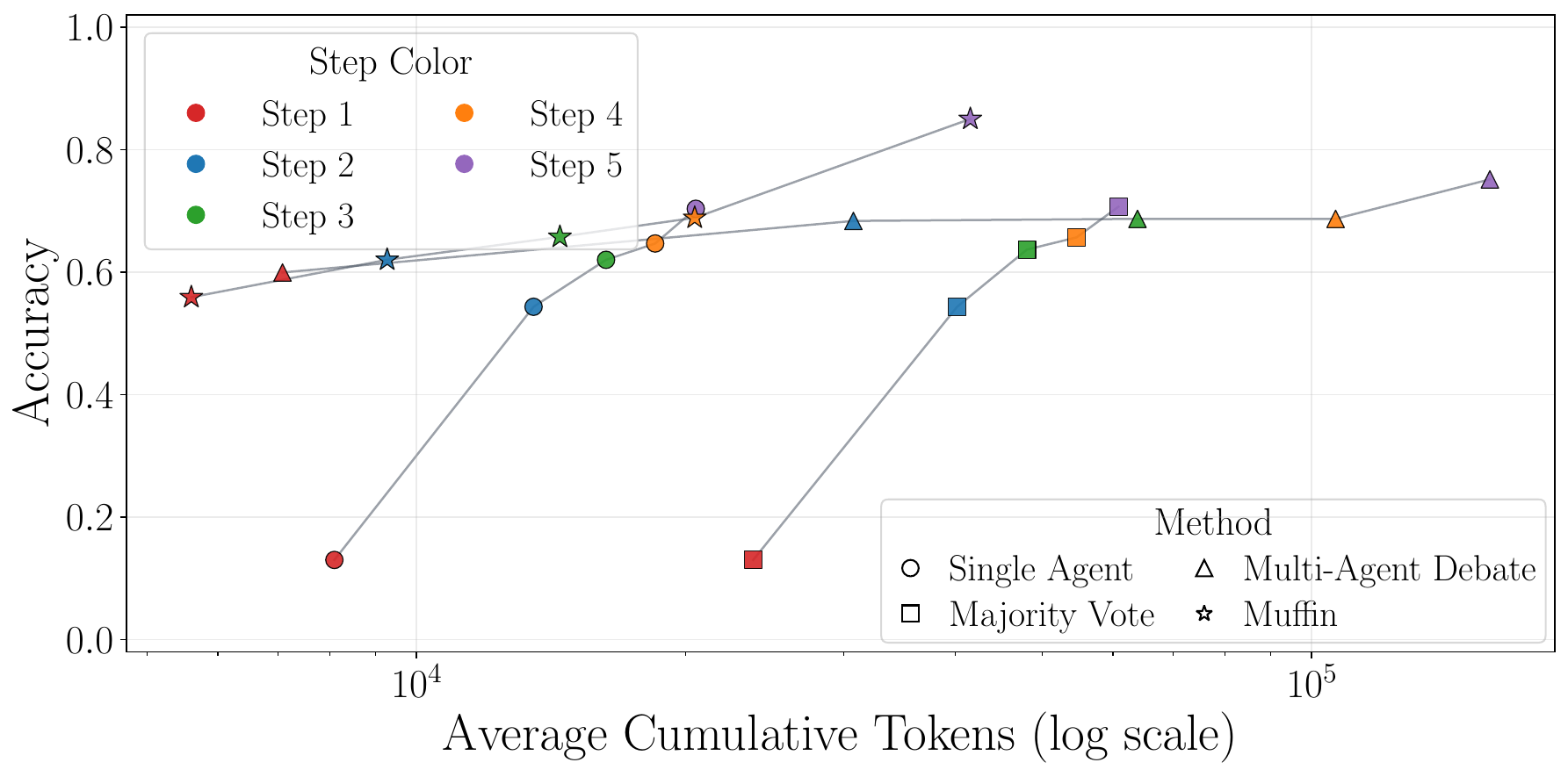}
        \caption{Step-wise accuracy and token usage on OlymMATH-Easy with \texttt{gemma-4-31B-it}.}
        \label{fig:app_step_token_olymmath_easy}
    \end{minipage}
\end{figure}

\begin{figure}[H]
    \begin{minipage}[t]{0.49\textwidth}
        \centering
        \includegraphics[width=1\linewidth]{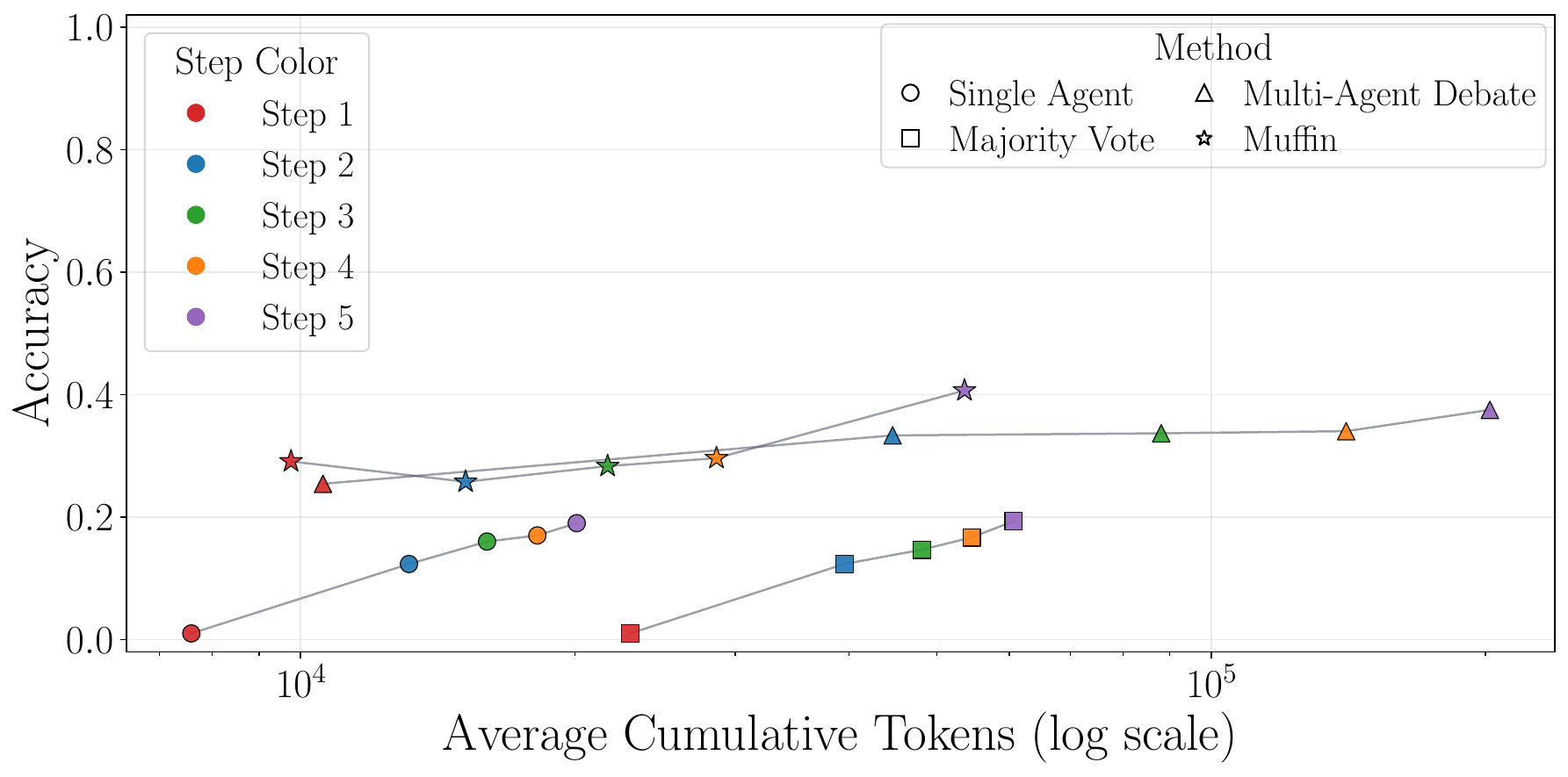}
        \caption{Step-wise accuracy and token usage on OlymMATH-Hard with \texttt{gemma-4-31B-it}.}
        \label{fig:app_step_token_olymmath_hard}
    \end{minipage}
    \hfill
    \begin{minipage}[t]{0.49\textwidth}
        \centering
        \includegraphics[width=1\linewidth]{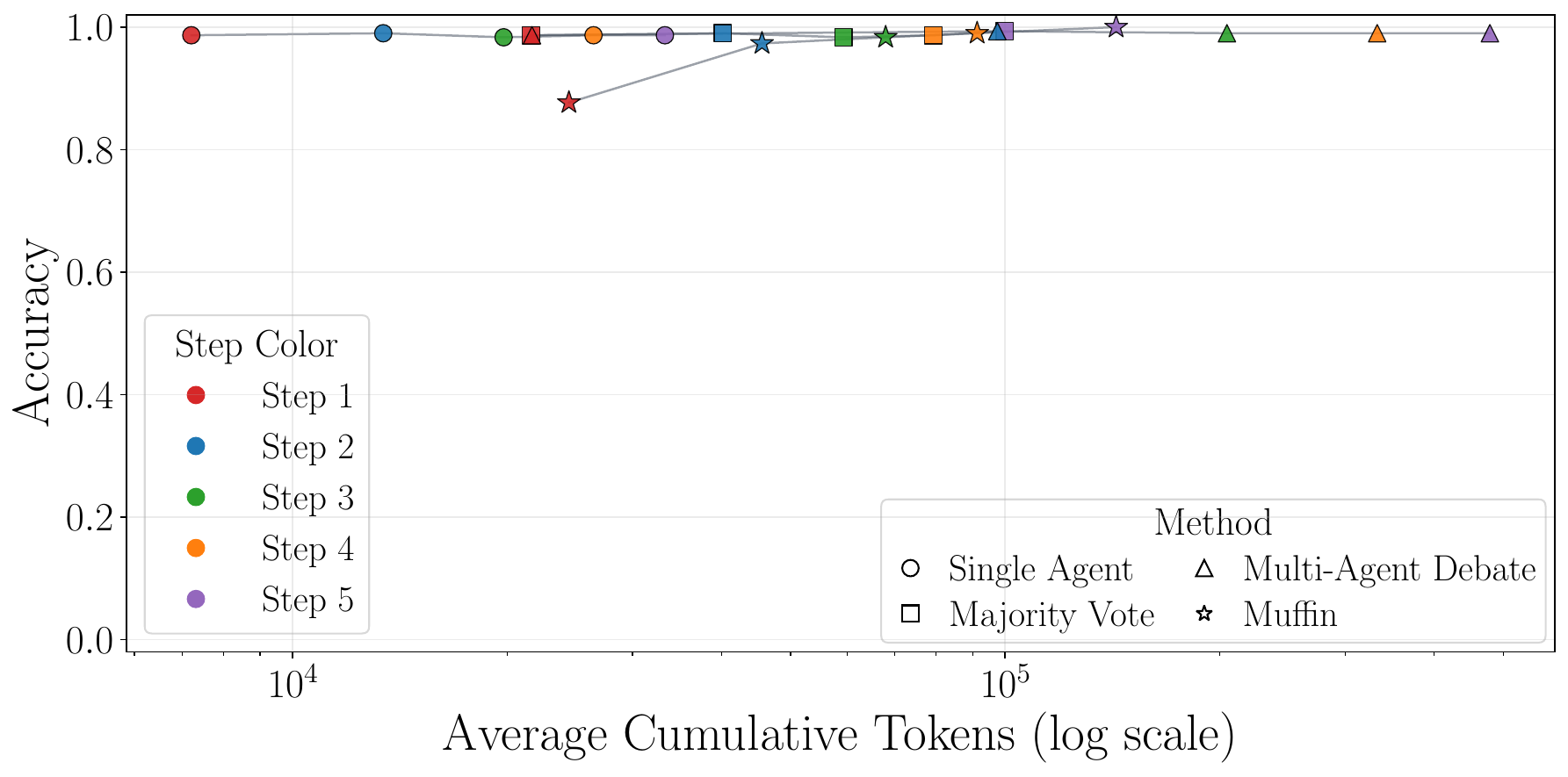}
        \caption{Step-wise accuracy and token usage on Chess with \texttt{Qwen3.8-27B}.}
        \label{fig:app_step_token_chess_qwen_27B}
    \end{minipage}
\end{figure}

\begin{figure}[H]
    \begin{minipage}[t]{0.49\textwidth}
        \centering
        \includegraphics[width=1\linewidth]{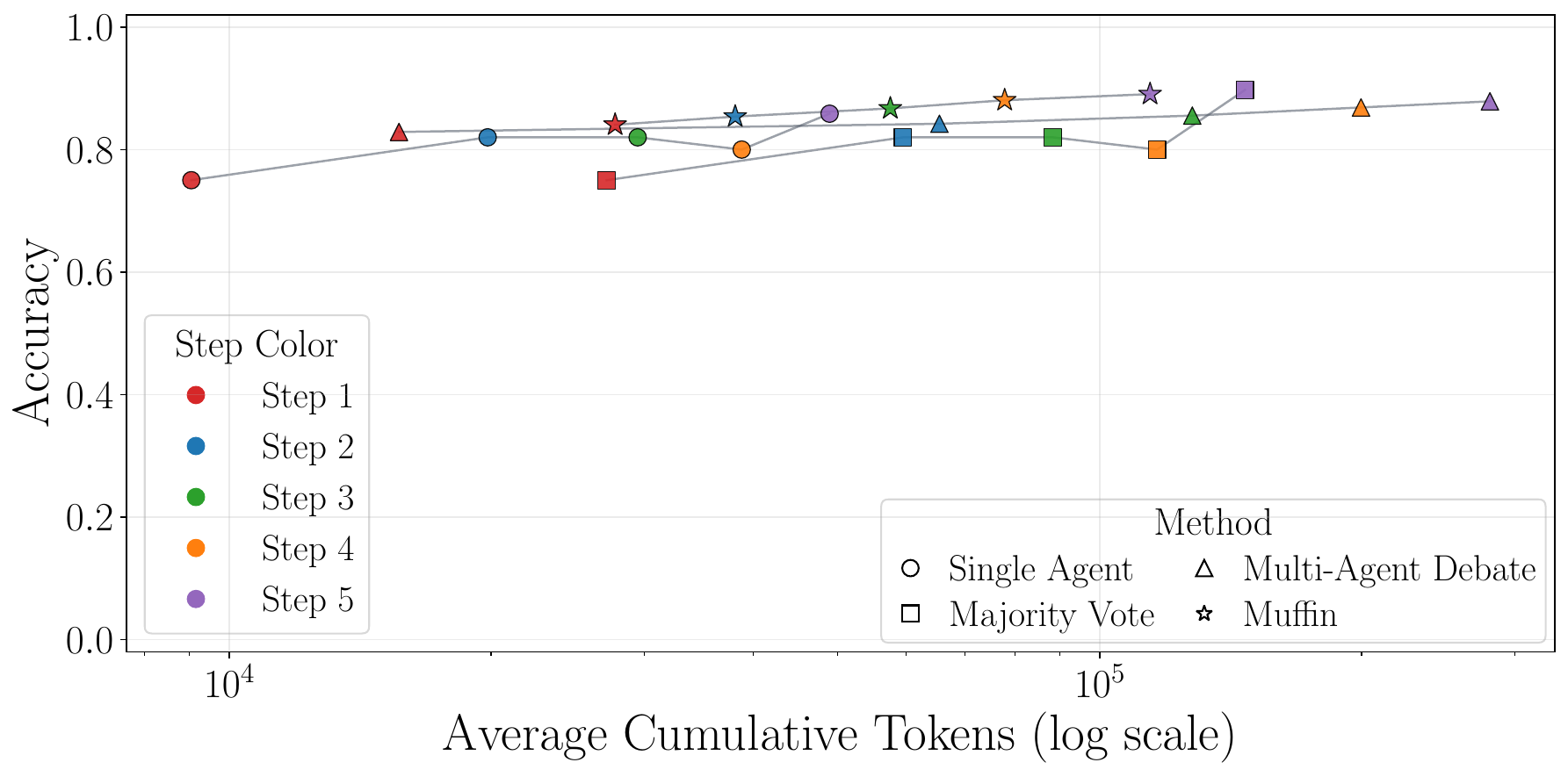}
        \caption{Step-wise accuracy and token usage on GPQA with \texttt{Qwen3.8-27B}.}
        \label{fig:app_step_token_gpqa_qwen_27B}
    \end{minipage}
    \hfill
    \begin{minipage}[t]{0.49\textwidth}
        \centering
        \includegraphics[width=1\linewidth]{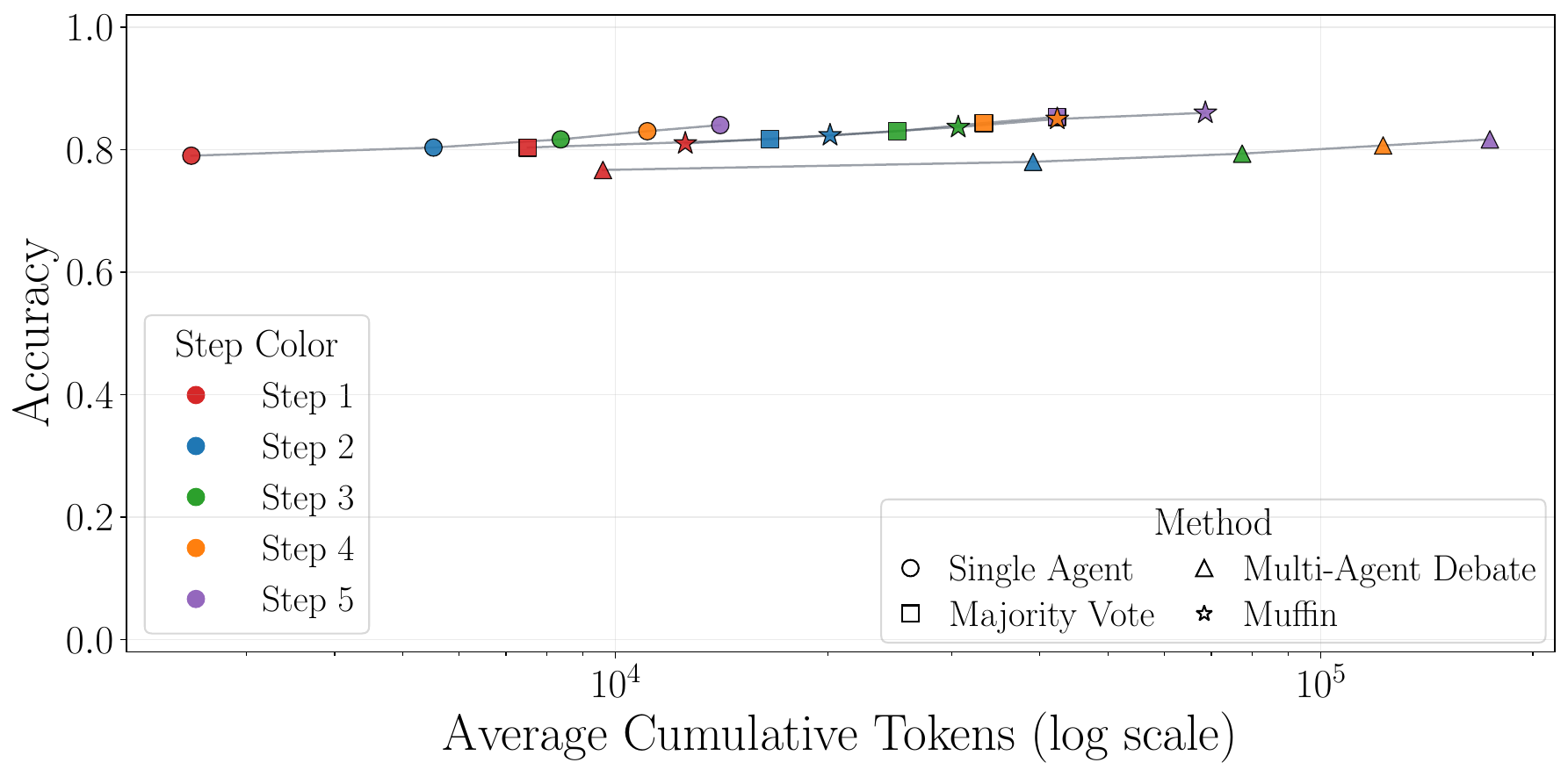}
        \caption{Step-wise accuracy and token usage on MMLU-Pro with \texttt{Qwen3.8-27B}.}
        \label{fig:app_step_token_mmlu_pro_qwen_27B}
    \end{minipage}
\end{figure}

\begin{figure}[H]
    \begin{minipage}[t]{0.49\textwidth}
        \centering
        \includegraphics[width=1\linewidth]{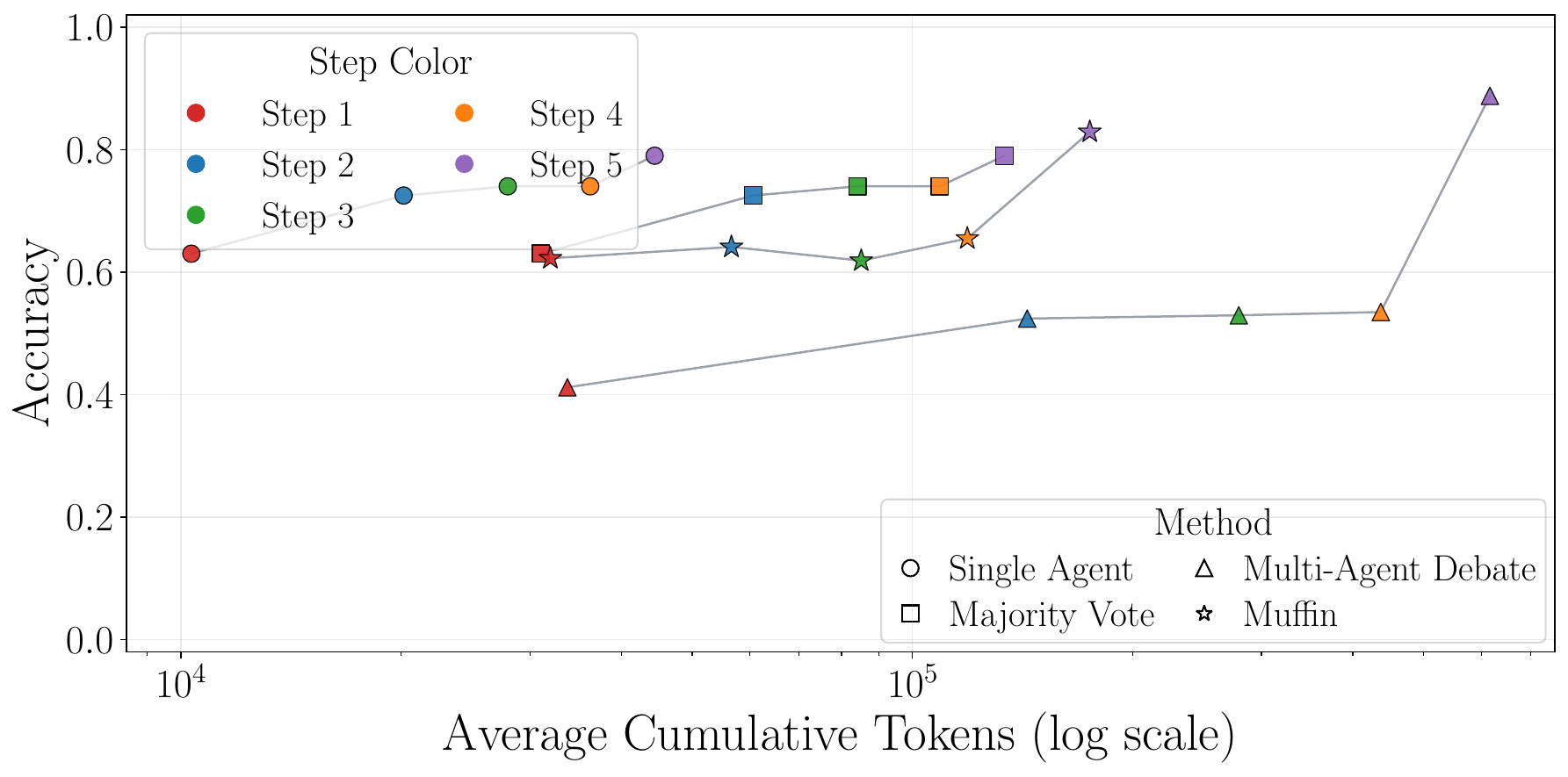}
        \caption{Step-wise accuracy and token usage on OlymMATH-Easy with \texttt{Qwen3.8-27B}.}
        \label{fig:app_step_token_olymmath_easy_qwen_27B}
    \end{minipage}
    \hfill
    \begin{minipage}[t]{0.49\textwidth}
        \centering
        \includegraphics[width=1\linewidth]{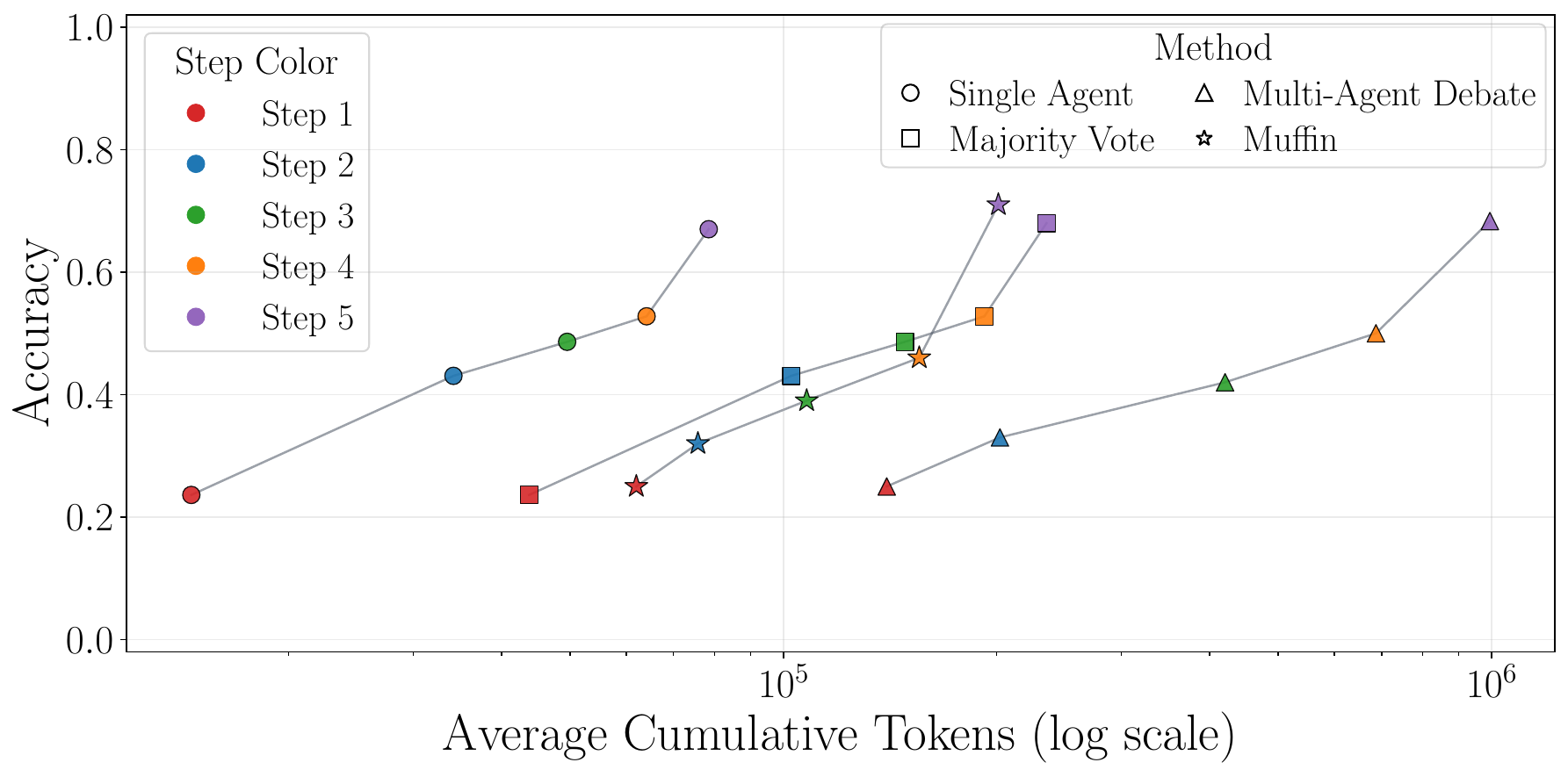}
        \caption{Step-wise accuracy and token usage on OlymMATH-Hard with \texttt{Qwen3.8-27B}.}
        \label{fig:app_step_token_olymmath_hard_qwen_27B}
    \end{minipage}
\end{figure}

Figures~\ref{fig:app_step_token_chess}--\ref{fig:app_step_token_olymmath_hard} and Figures~\ref{fig:app_step_token_chess_qwen_27B}--\ref{fig:app_step_token_olymmath_hard_qwen_27B} show the 5-step accuracy and token usage across all benchmarks using \texttt{gemma-4-31B-it} and \texttt{Qwen3.8-27B}, respectively.

\subsection{Shared findings insights}
\label{app:shared_findings}

\begin{figure}[H]
    \begin{minipage}[t]{0.49\textwidth}
        \centering
        \includegraphics[width=\linewidth]{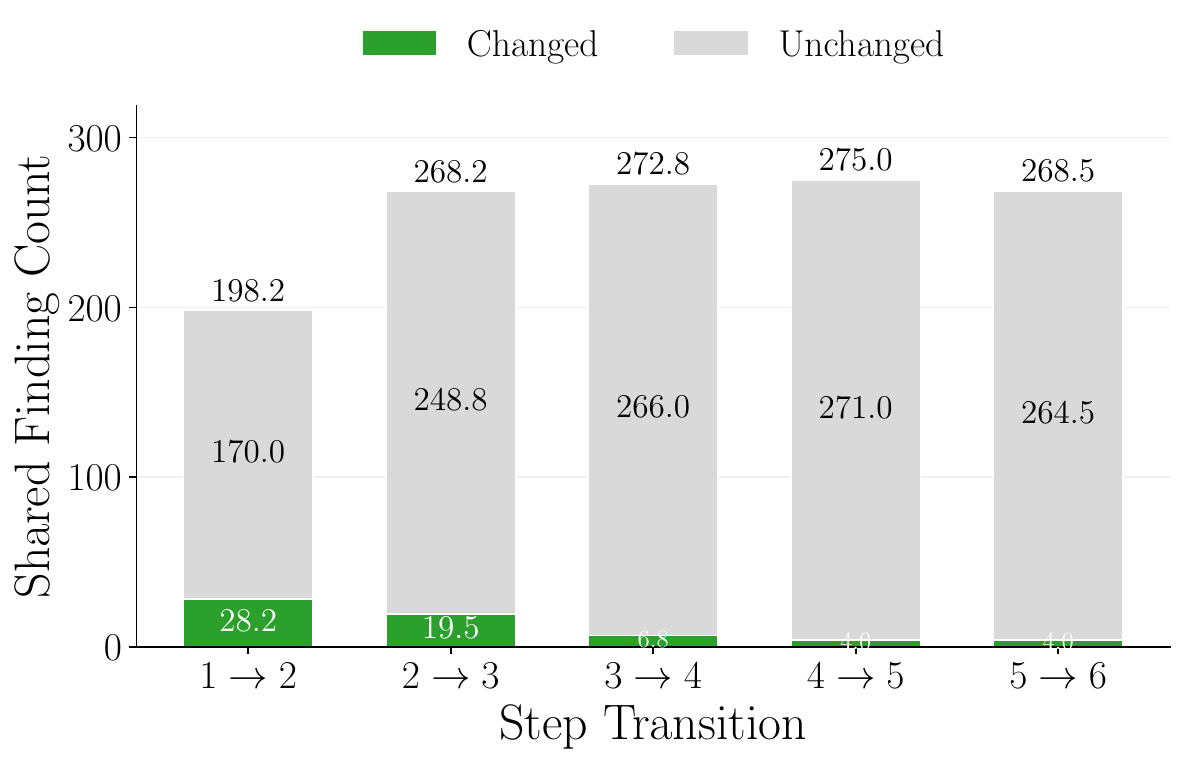}
        \caption{Mean number of shared findings that change or do not change an agent's answer on GPQA with \texttt{gemma-4-31B-it}.}
        \label{fig:app_change_gpqa}
    \end{minipage}
    \hfill
    \begin{minipage}[t]{0.49\textwidth}
        \centering
        \includegraphics[width=\linewidth]{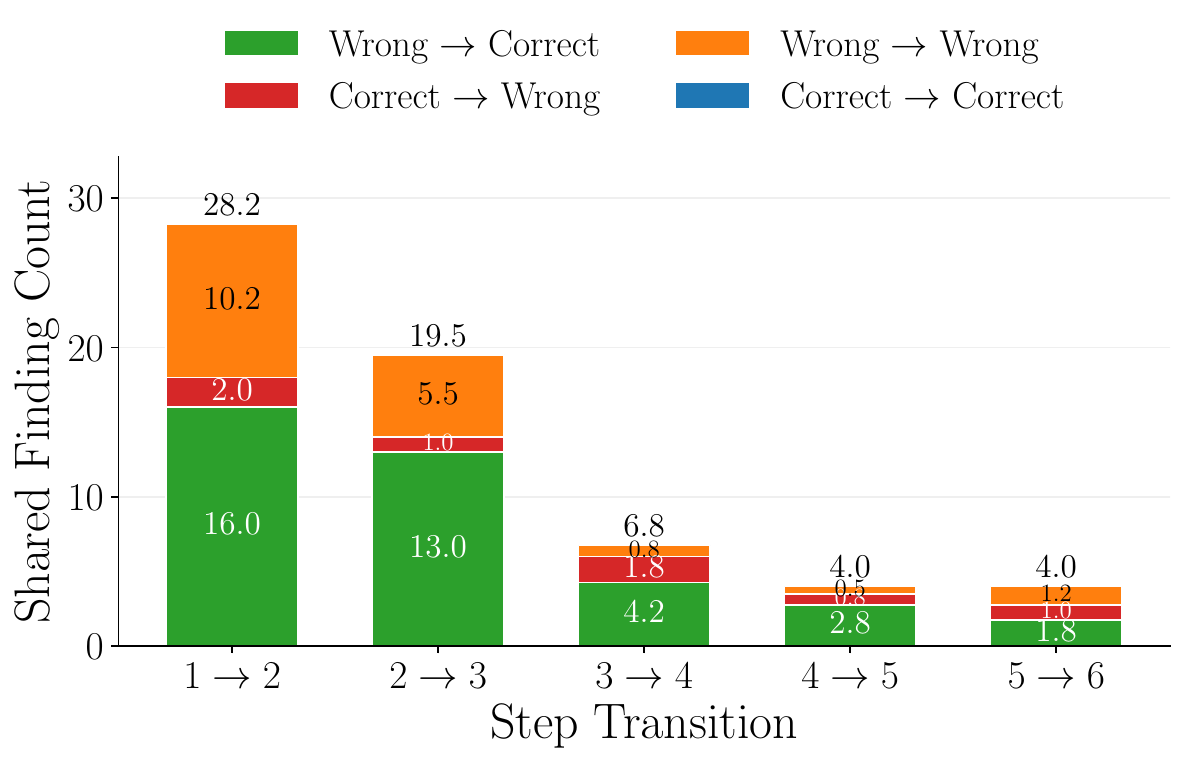}
        \caption{Outcome breakdown for shared findings that change an agent's answer on GPQA with \texttt{gemma-4-31B-it}.}
        \label{fig:app_change_gpqa_detailed}
    \end{minipage}
\end{figure}

\begin{figure}[H]
    \begin{minipage}[t]{0.49\textwidth}
        \centering
        \includegraphics[width=\linewidth]{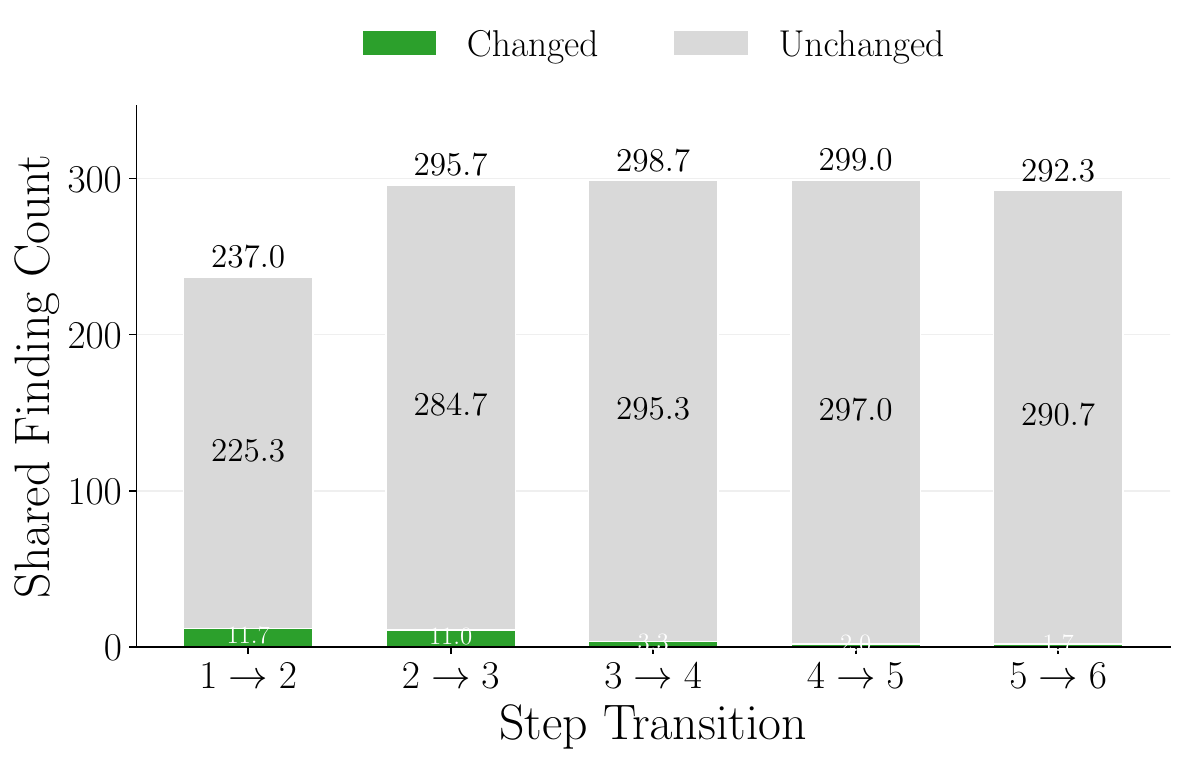}
        \caption{Mean number of shared findings that change or do not change an agent's answer on MMLU-Pro with \texttt{gemma-4-31B-it}.}
        \label{fig:app_change_mmlu_pro}
    \end{minipage}
    \hfill
    \begin{minipage}[t]{0.49\textwidth}
        \centering
        \includegraphics[width=\linewidth]{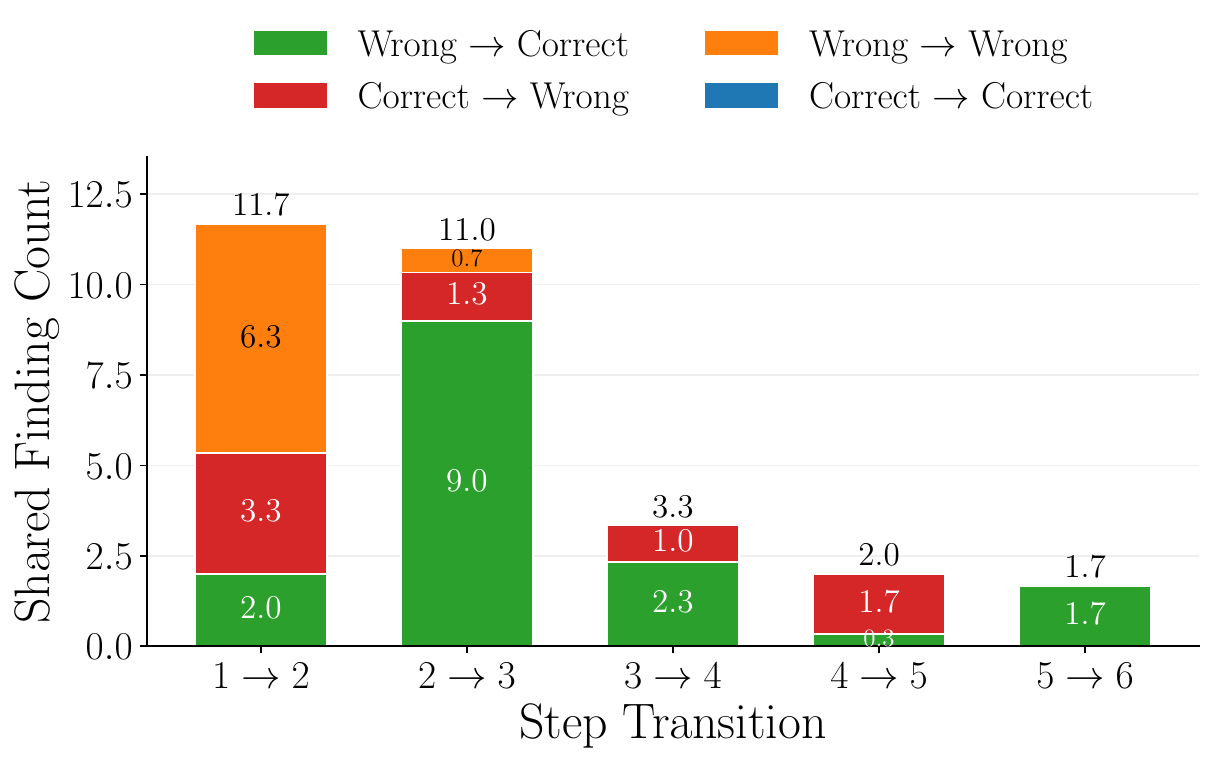}
        \caption{Outcome breakdown for shared findings that change an agent's answer on MMLU-Pro with \texttt{gemma-4-31B-it}.}
        \label{fig:app_change_mmlu_pro_detailed}
    \end{minipage}
\end{figure}

\begin{figure}[H]
    \begin{minipage}[t]{0.49\textwidth}
        \centering
        \includegraphics[width=\linewidth]{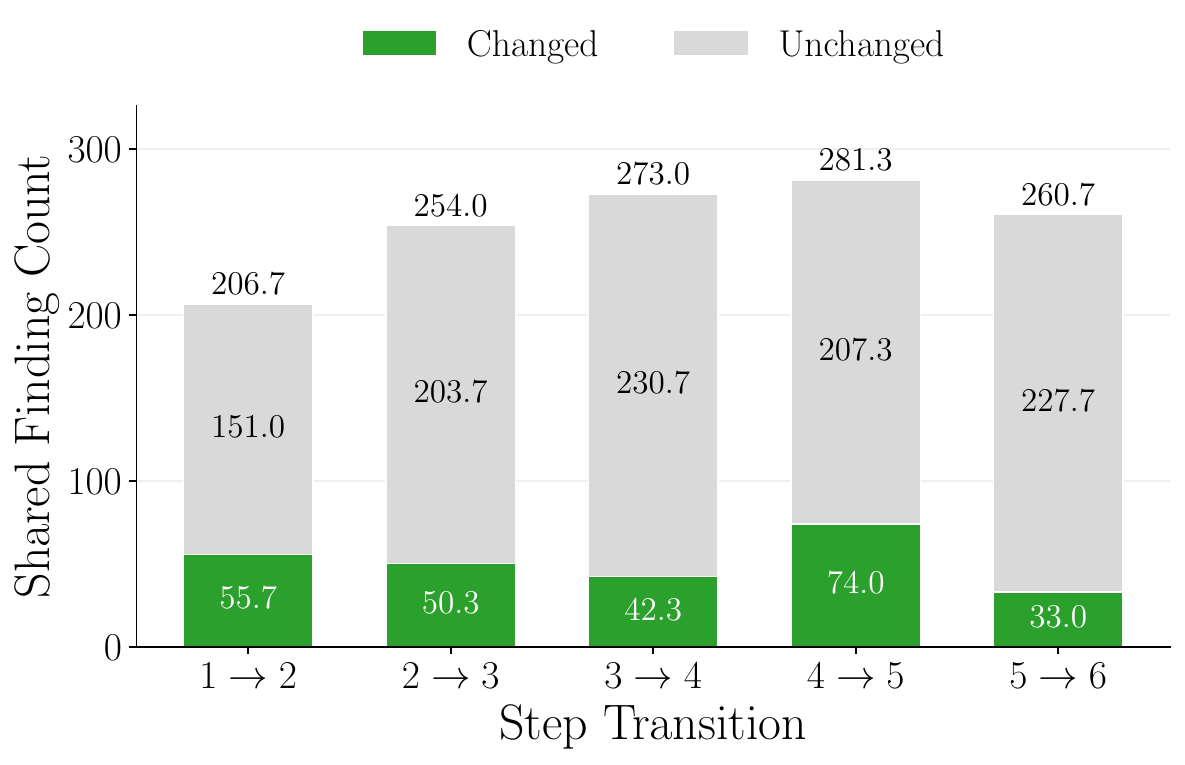}
        \caption{Mean number of shared findings that change or do not change an agent's answer on OlymMATH-Easy with \texttt{gemma-4-31B-it}.}
        \label{fig:app_change_olymmath_easy}
    \end{minipage}
    \hfill
    \begin{minipage}[t]{0.49\textwidth}
        \centering
        \includegraphics[width=\linewidth]{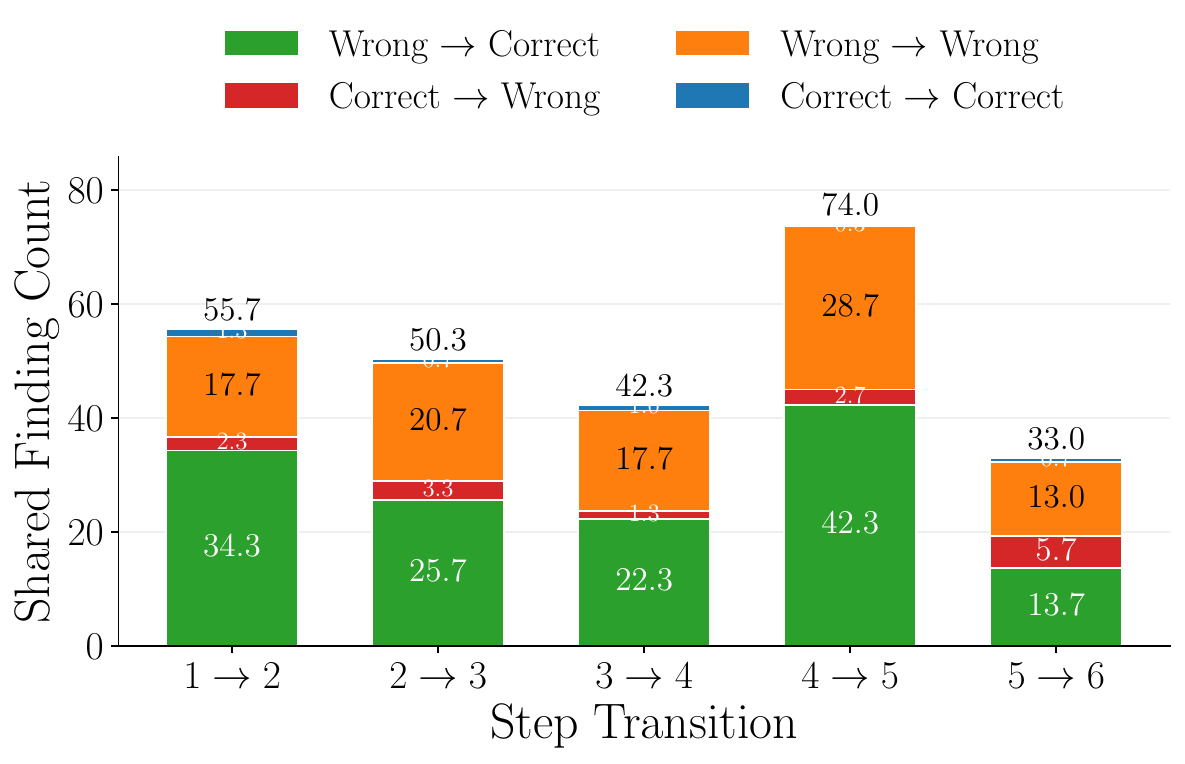}
        \caption{Outcome breakdown for shared findings that change an agent's answer on OlymMATH-Easy with \texttt{gemma-4-31B-it}.}
        \label{fig:app_change_olymmath_easy_detailed}
    \end{minipage}
\end{figure}

\begin{figure}[H]
    \begin{minipage}[t]{0.49\textwidth}
        \centering
        \includegraphics[width=\linewidth]{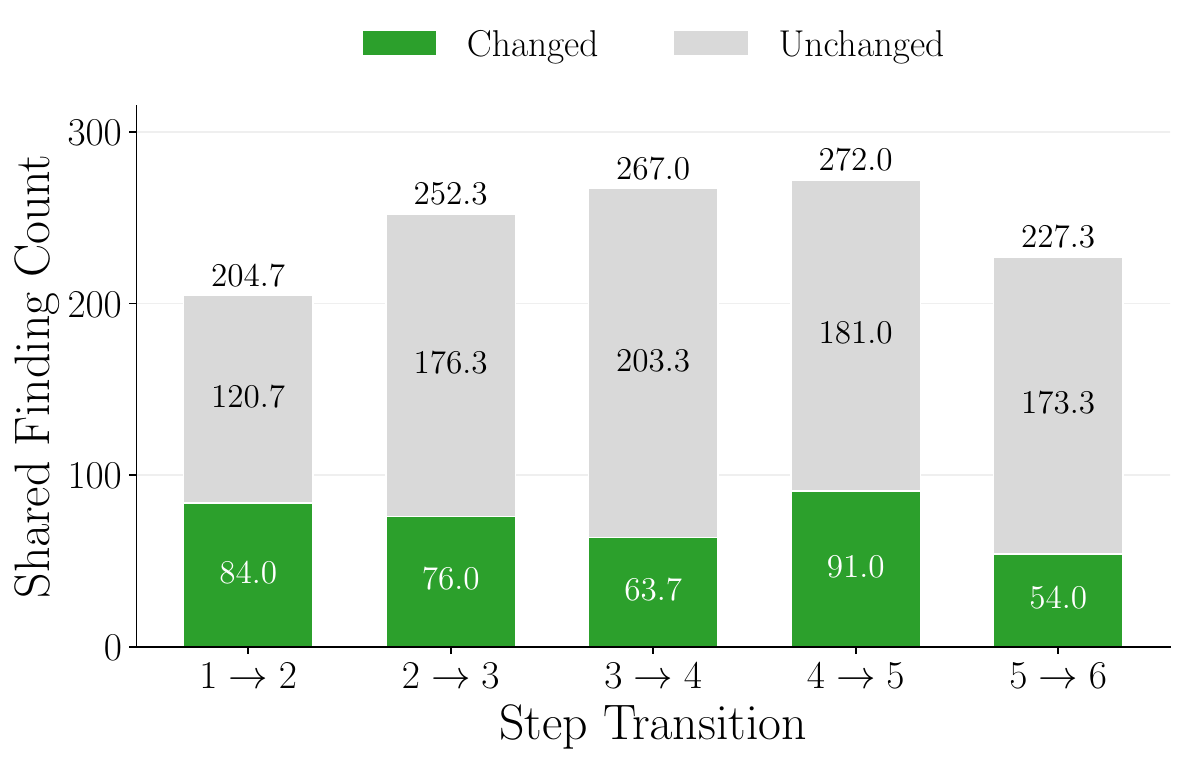}
        \caption{Mean number of shared findings that change or do not change an agent's answer on OlymMATH-Hard with \texttt{gemma-4-31B-it}.}
        \label{fig:app_change_olymmath_hard}
    \end{minipage}
    \hfill
    \begin{minipage}[t]{0.49\textwidth}
        \centering
        \includegraphics[width=\linewidth]{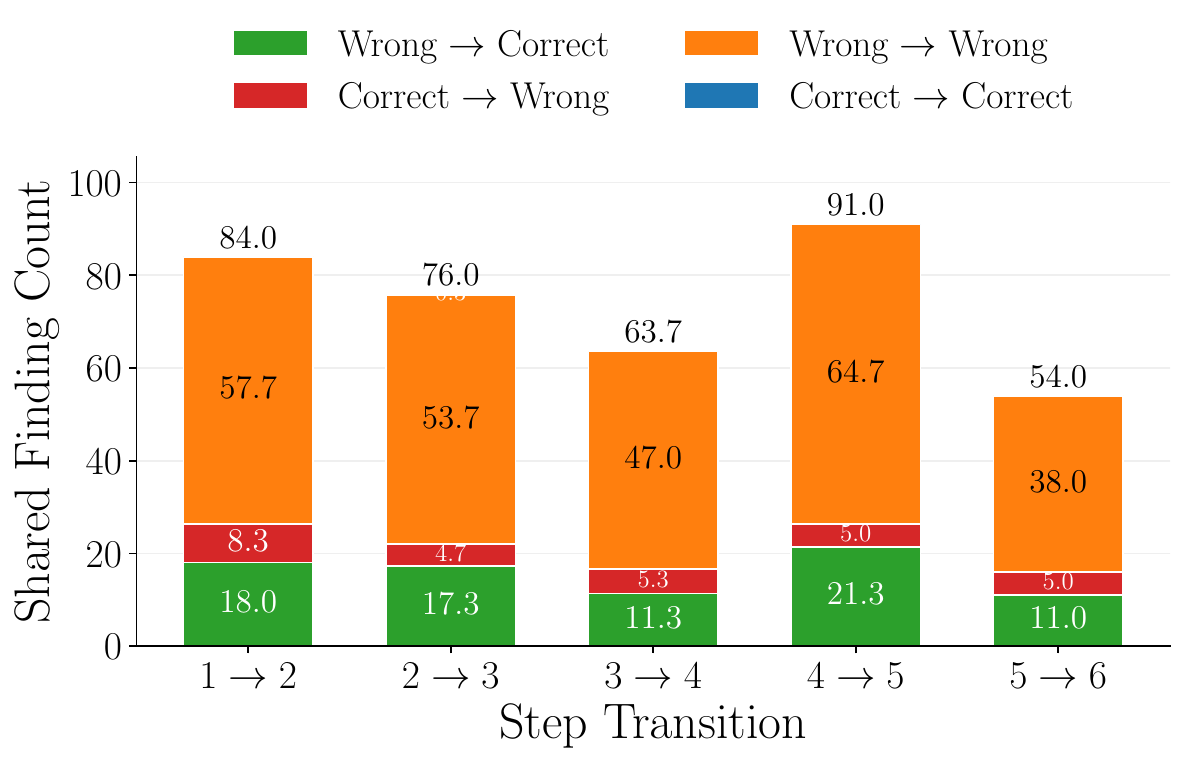}
        \caption{Outcome breakdown for shared findings that change an agent's answer on OlymMATH-Hard with \texttt{gemma-4-31B-it}.}
        \label{fig:app_change_olymmath_hard_detailed}
    \end{minipage}
\end{figure}

\begin{figure}[H]
    \begin{minipage}[t]{0.49\textwidth}
        \centering
        \includegraphics[width=\linewidth]{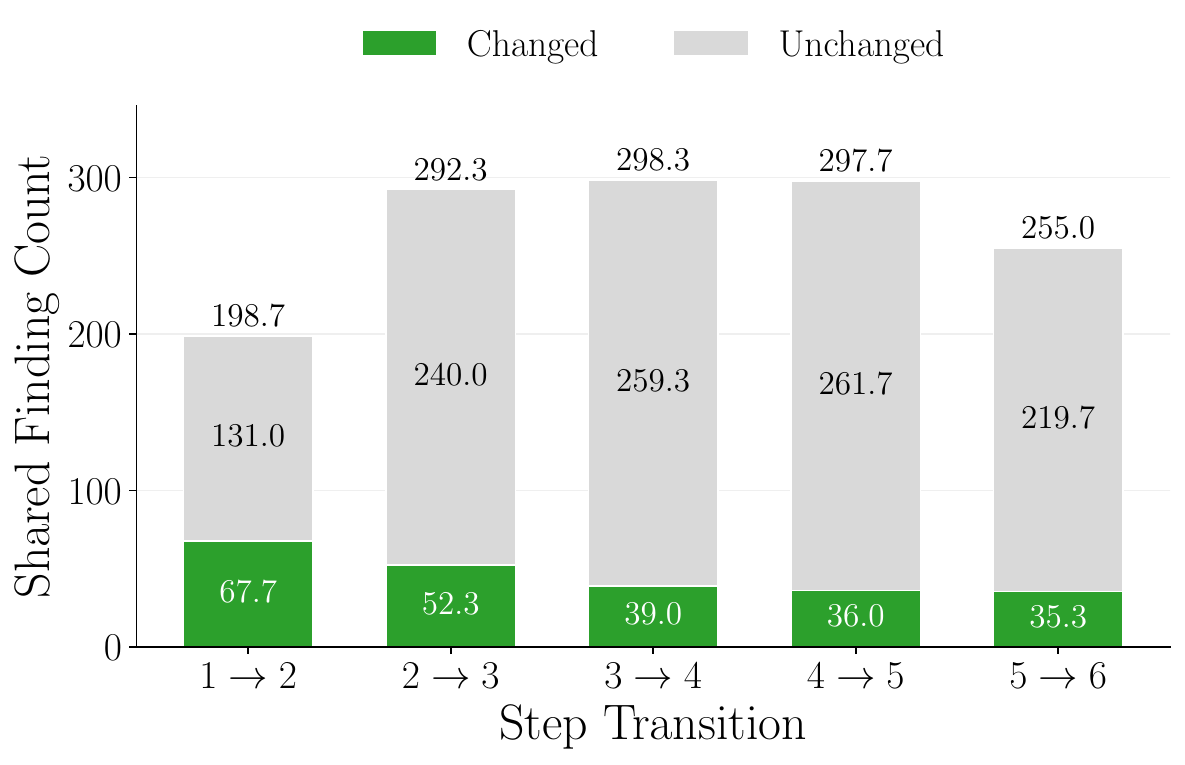}
        \caption{Mean number of shared findings that change or do not change an agent's answer on Chess with \texttt{Qwen3.8-27B}.}
        \label{fig:app_change_chess_27b}
    \end{minipage}
    \hfill
    \begin{minipage}[t]{0.49\textwidth}
        \centering
        \includegraphics[width=\linewidth]{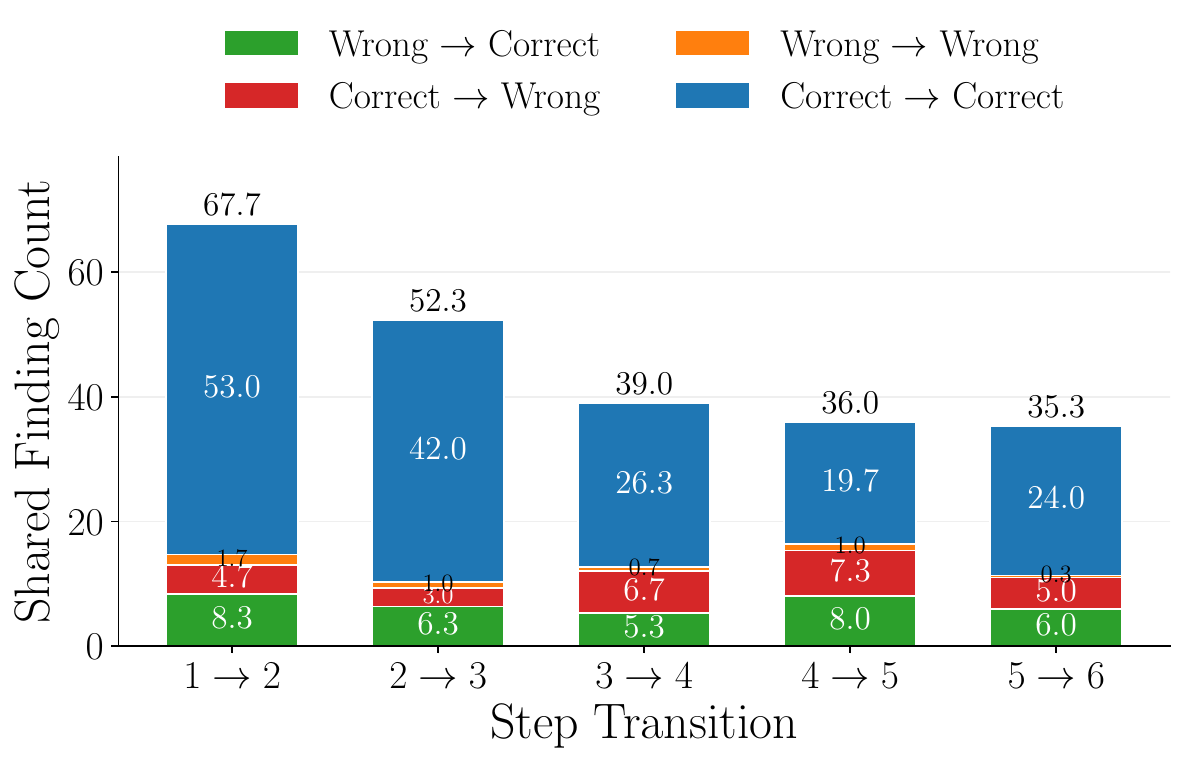}
        \caption{Outcome breakdown for shared findings that change an agent's answer on Chess with \texttt{Qwen3.8-27B}.}
        \label{fig:app_change_chess_detailed_27b}
    \end{minipage}
\end{figure}

\begin{figure}[H]
    \begin{minipage}[t]{0.49\textwidth}
        \centering
        \includegraphics[width=\linewidth]{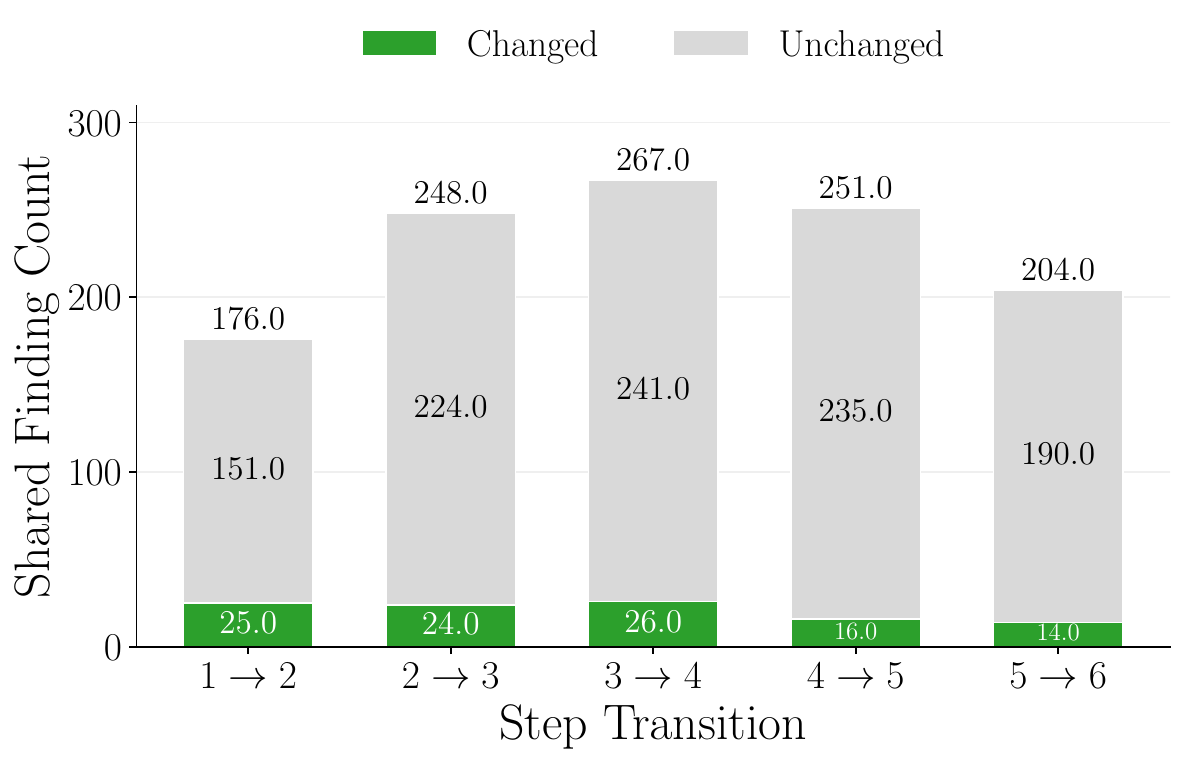}
        \caption{Mean number of shared findings that change or do not change an agent's answer on GPQA with \texttt{Qwen3.8-27B}.}
        \label{fig:app_change_gpqa_27b}
    \end{minipage}
    \hfill
    \begin{minipage}[t]{0.49\textwidth}
        \centering
        \includegraphics[width=\linewidth]{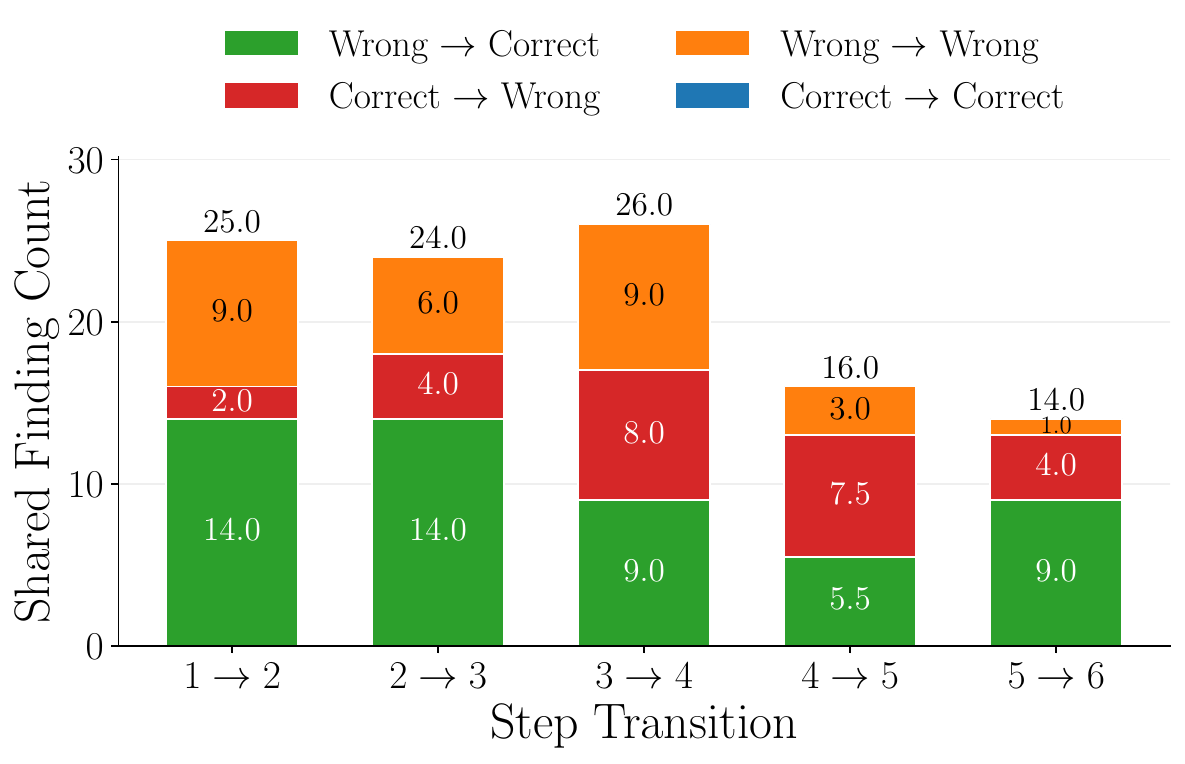}
        \caption{Outcome breakdown for shared findings that change an agent's answer on GPQA with \texttt{Qwen3.8-27B}.}
        \label{fig:app_change_gpqa_detailed_27b}
    \end{minipage}
\end{figure}

\begin{figure}[H]
    \begin{minipage}[t]{0.49\textwidth}
        \centering
        \includegraphics[width=\linewidth]{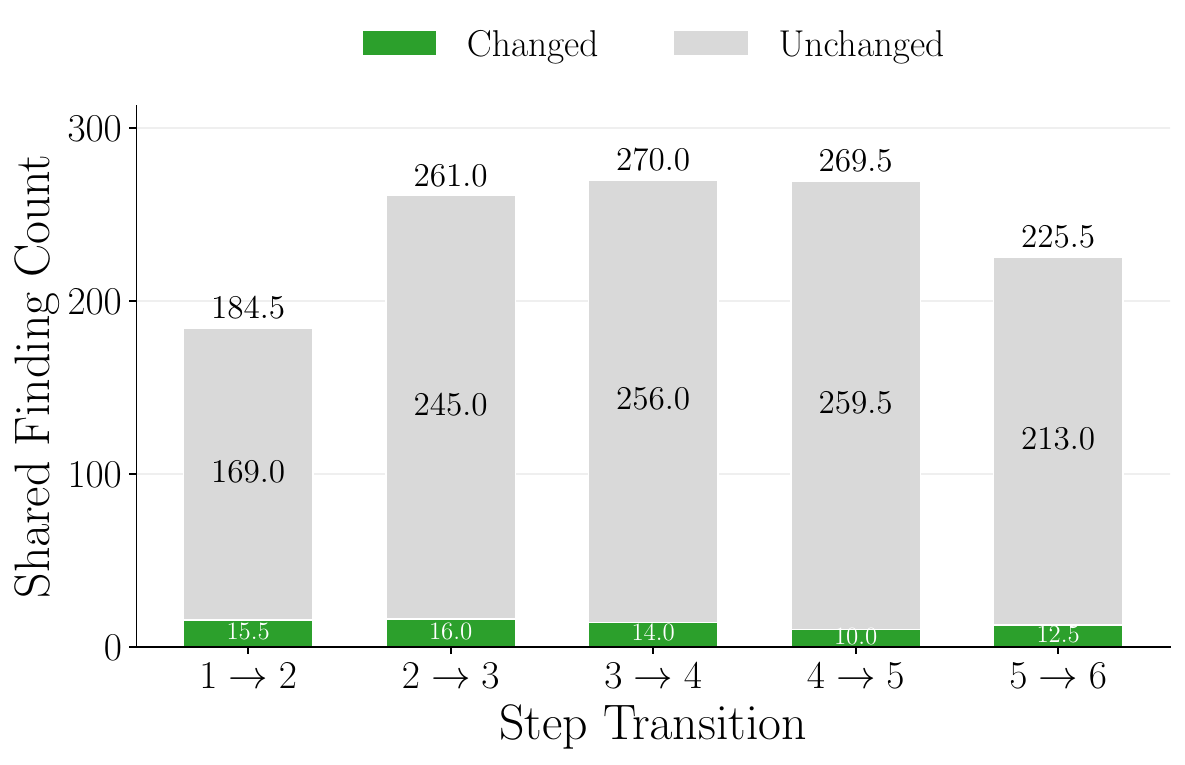}
        \caption{Mean number of shared findings that change or do not change an agent's answer on MMLU-Pro with \texttt{Qwen3.8-27B}.}
        \label{fig:app_change_mmlu_pro_27b}
    \end{minipage}
    \hfill
    \begin{minipage}[t]{0.49\textwidth}
        \centering
        \includegraphics[width=\linewidth]{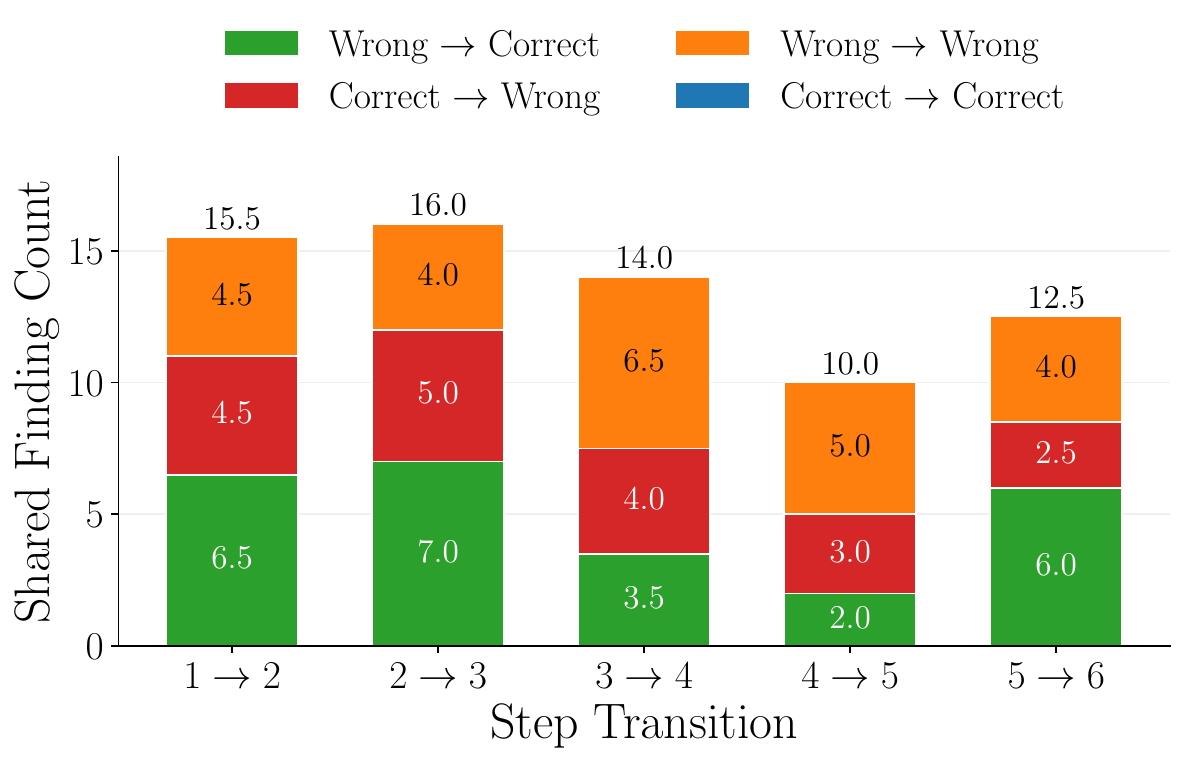}
        \caption{Outcome breakdown for shared findings that change an agent's answer on MMLU-Pro with \texttt{Qwen3.8-27B}.}
        \label{fig:app_change_mmlu_pro_detailed_27b}
    \end{minipage}
\end{figure}

\begin{figure}[H]
    \begin{minipage}[t]{0.49\textwidth}
        \centering
        \includegraphics[width=\linewidth]{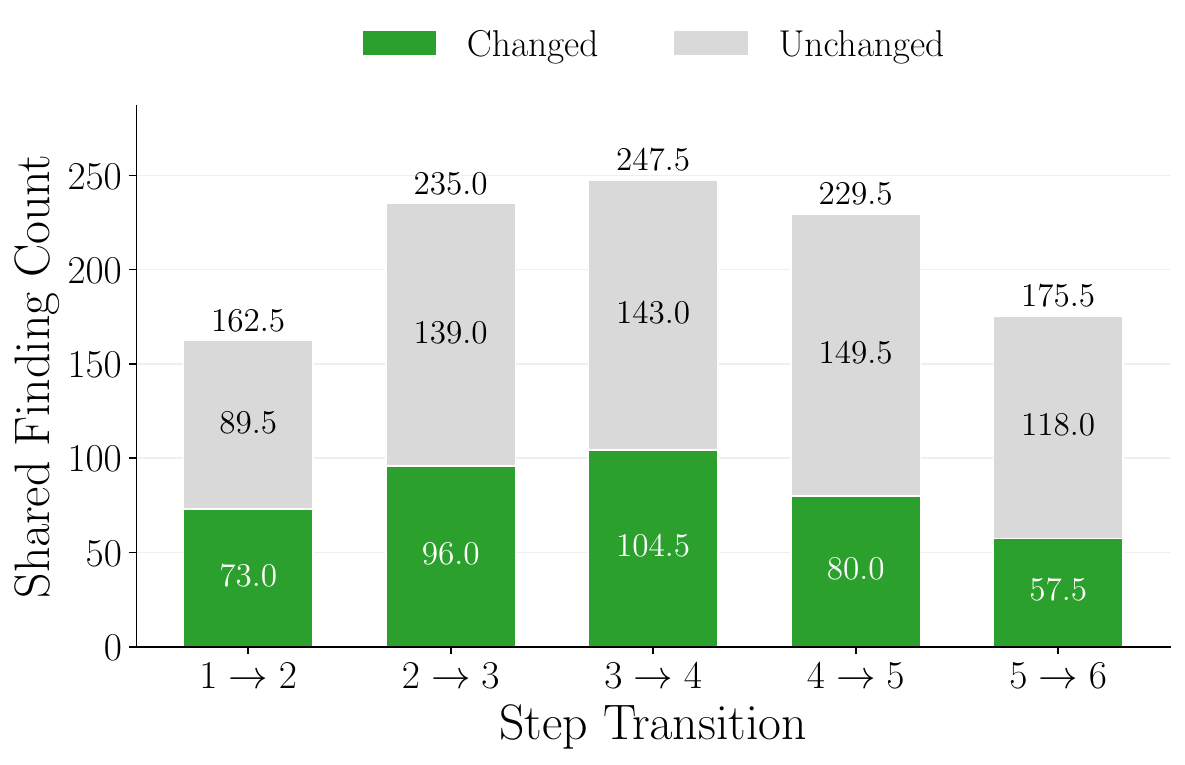}
        \caption{Mean number of shared findings that change or do not change an agent's answer on OlymMATH-Easy with \texttt{Qwen3.8-27B}.}
        \label{fig:app_change_olymmath_easy_27b}
    \end{minipage}
    \hfill
    \begin{minipage}[t]{0.49\textwidth}
        \centering
        \includegraphics[width=\linewidth]{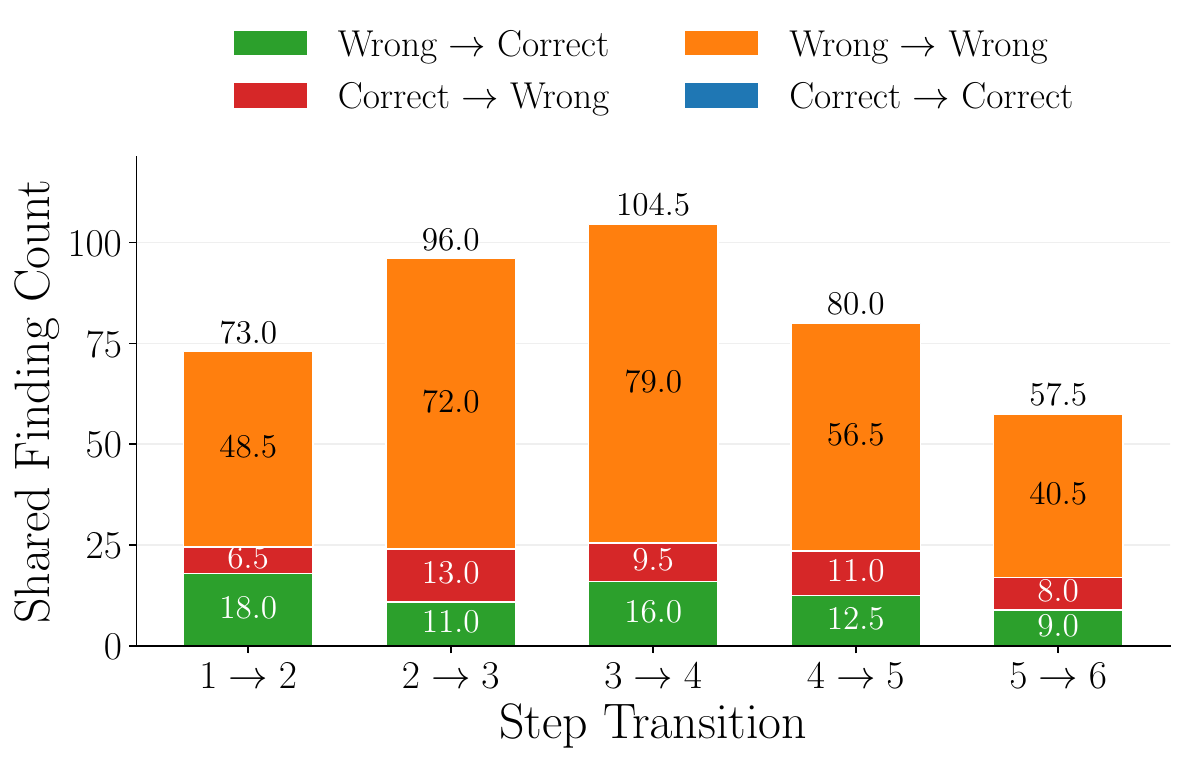}
        \caption{Outcome breakdown for shared findings that change an agent's answer on OlymMATH-Easy with \texttt{Qwen3.8-27B}.}
        \label{fig:app_change_olymmath_easy_detailed_27b}
    \end{minipage}
\end{figure}

\begin{figure}[H]
    \begin{minipage}[t]{0.49\textwidth}
        \centering
        \includegraphics[width=\linewidth]{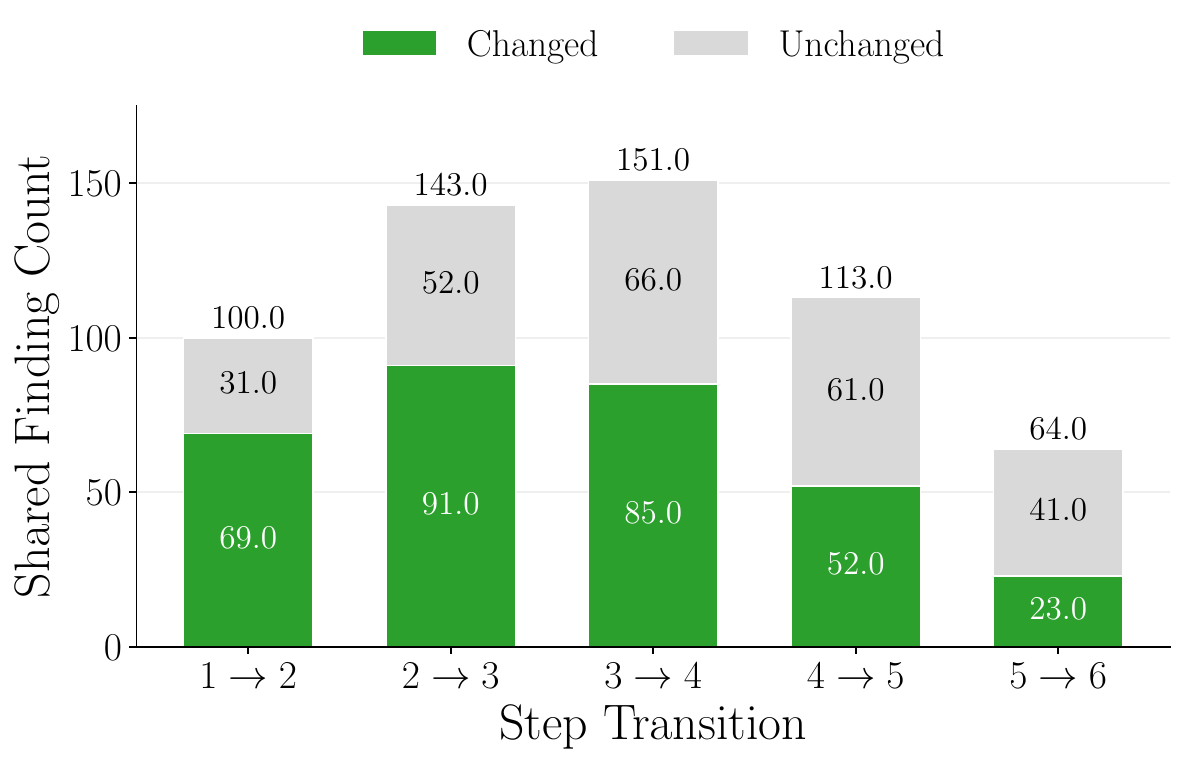}
        \caption{Mean number of shared findings that change or do not change an agent's answer on OlymMATH-Hard with \texttt{Qwen3.8-27B}.}
        \label{fig:app_change_olymmath_hard_27b}
    \end{minipage}
    \hfill
    \begin{minipage}[t]{0.49\textwidth}
        \centering
        \includegraphics[width=\linewidth]{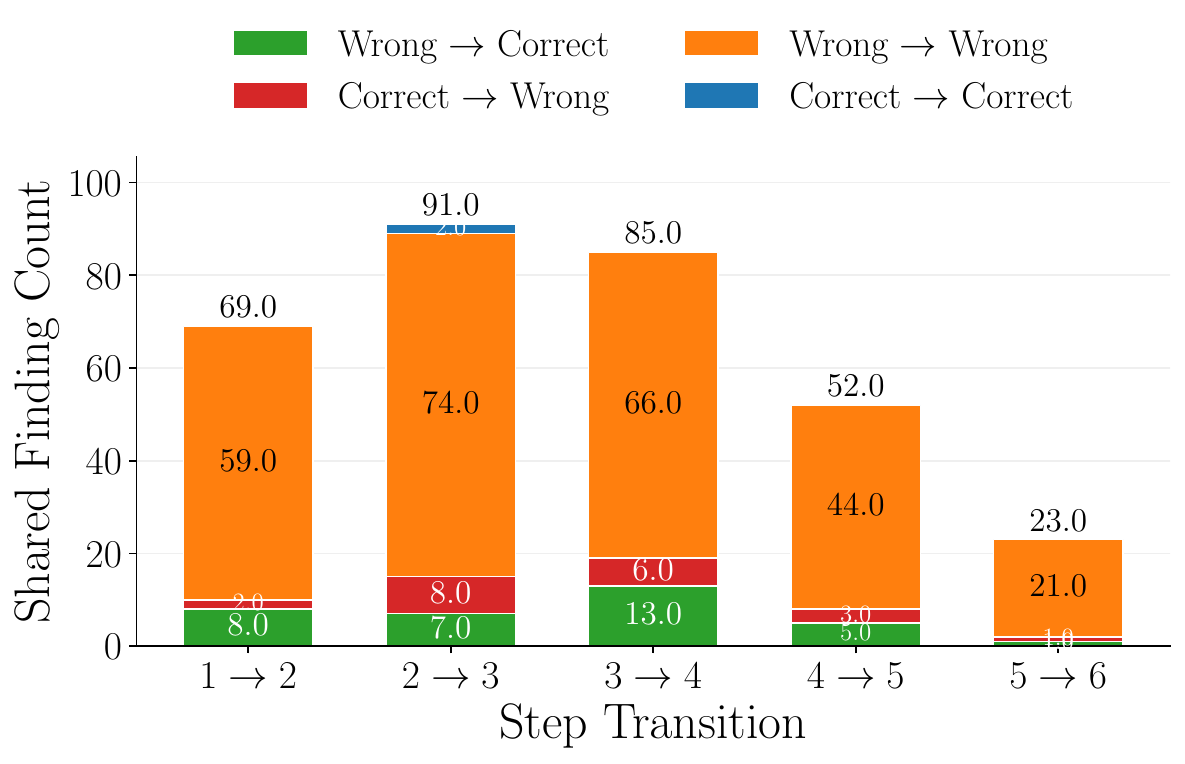}
        \caption{Outcome breakdown for shared findings that change an agent's answer on OlymMATH-Hard with \texttt{Qwen3.8-27B}.}
        \label{fig:app_change_olymmath_hard_detailed_27b}
    \end{minipage}
\end{figure}

Figures~\ref{fig:app_change_gpqa}--\ref{fig:app_change_olymmath_hard_detailed} show the shared-finding answer-change analysis for the remaining benchmarks with \texttt{gemma-4-31B-it}, while Figures~\ref{fig:app_change_chess_27b}--\ref{fig:app_change_olymmath_hard_detailed_27b} provide the corresponding results for \texttt{Qwen3.8-27B}.

